\documentclass[11pt]{article}

\usepackage{arxiv}

\usepackage[T1]{fontenc}
\usepackage[utf8]{inputenc}

\usepackage{amsmath,amssymb}
\usepackage{array}
\usepackage{booktabs}
\usepackage{multirow}
\usepackage{tabularx}

\usepackage{graphicx}
\usepackage{subfig}
\usepackage{float}

\usepackage{algorithm}
\usepackage{algpseudocode}

\usepackage[numbers]{natbib}

\usepackage{xcolor}
\usepackage{url}
\usepackage{hyperref}
\usepackage{cleveref}

\graphicspath{{.}}

\usepackage[acronym, nohypertypes={acronym}]{glossaries}
\glsdisablehyper

\newacronym{sar}{SAR}{Synthetic Aperture Radar}
\newacronym{dem}{DEM}{Digital Elevation Map}
\newacronym{met}{Met}{Meterological fields}

\newacronym{vit}{ViT}{Vision Transformer}
\newacronym{cnn}{CNN}{Convolutional Neural Network}
\newacronym{sam}{SAM}{Segment Anything Model}
\newacronym{samm}{SAMM}{SAM with Multiple Modalities}
\newacronym{sfg}{SFG}{Selective Fusion Gate}
\newacronym{mlp}{MLP}{Multi-Layer Perceptron}
\newacronym{mae}{MAE}{Masked Autoencoder}

\newacronym{peft}{PEFT}{Parameter-Efficient Fine-Tuning}
\newacronym{lora}{LoRA}{Low-Rank Adaptation}
\newacronym{iou}{IoU}{Intersection over Union}
\newacronym{bb}{BB}{bounding box}
\newacronym{sa}{SA}{Slope Angle}
\newacronym{mse}{MSE}{Mean Squared Error}
\newglossaryentry{gpu}{name=NVIDIA RTX 6000 Ada,description=}
\newcommand{\PaperAbstract}{%
Remote sensing solutions for avalanche segmentation and mapping are key to supporting risk forecasting and mitigation in mountain regions. 
\gls{sar} imagery from Sentinel-1 can be effectively used for this task, but training an effective detection model requires gathering a large dataset with high-quality annotations from domain experts, which is prohibitively time-consuming. 
In this work, we aim to facilitate and accelerate the annotation of \gls{sar} images for avalanche mapping. 
We build on the \gls{sam}, a segmentation foundation model trained on natural images, and tailor it to Sentinel-1 \gls{sar} data. 
Adapting \gls{sam} to our use-case requires addressing several domain-specific challenges: 
(i) domain mismatch, since \gls{sam} was not trained on satellite/\gls{sar} imagery; 
(ii) input adaptation, because \gls{sar} products typically provide more than three channels, while \gls{sam} is constrained to RGB images; 
(iii) robustness to imprecise prompts that can affect target identification and degrade the segmentation quality, an issue exacerbated in small, low-contrast avalanches; 
and (iv) training efficiency, since standard fine-tuning is computationally demanding for \gls{sam}. 
We tackle these challenges through a combination of adapters to mitigate the domain gap, multiple encoders to handle multi-channel \gls{sar} inputs, prompt-engineering strategies to improve avalanche localization accuracy, and a training algorithm that limits the training time of the encoder, which is recognized as the major bottleneck. 
We integrate the resulting model into an annotation tool and show experimentally that it speeds up the annotation of \gls{sar} images. 
}

\newcommand{\PaperBody}{%
\section{Introduction}
\begin{figure}[!ht]
    \centering 
    
    \subfloat[\gls{sar}  Image 1]{\includegraphics[width=0.4\textwidth]{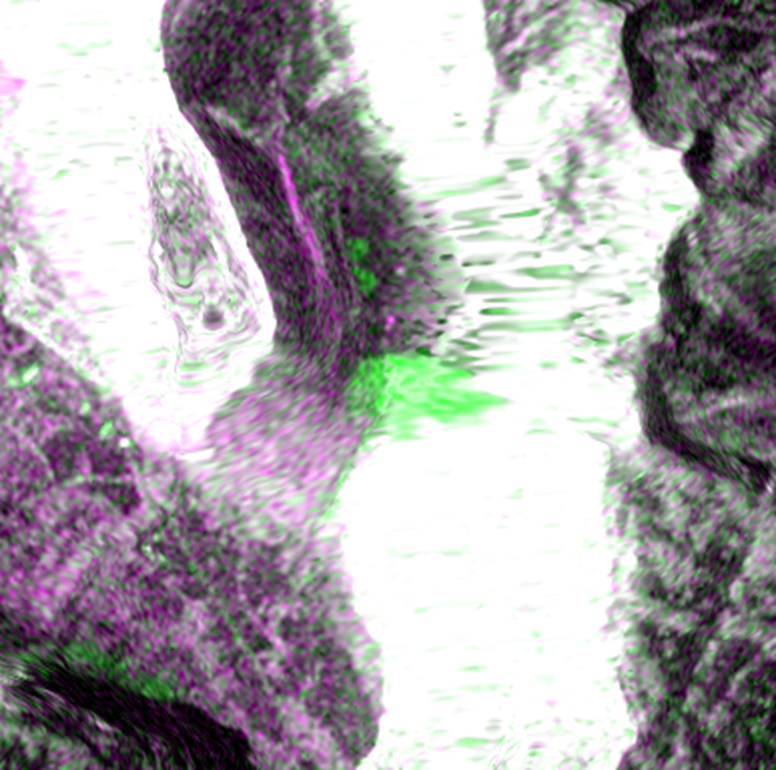}}
    \hspace{0.1cm}
    \subfloat[Ground Truth 1]{\includegraphics[width=0.4\textwidth]{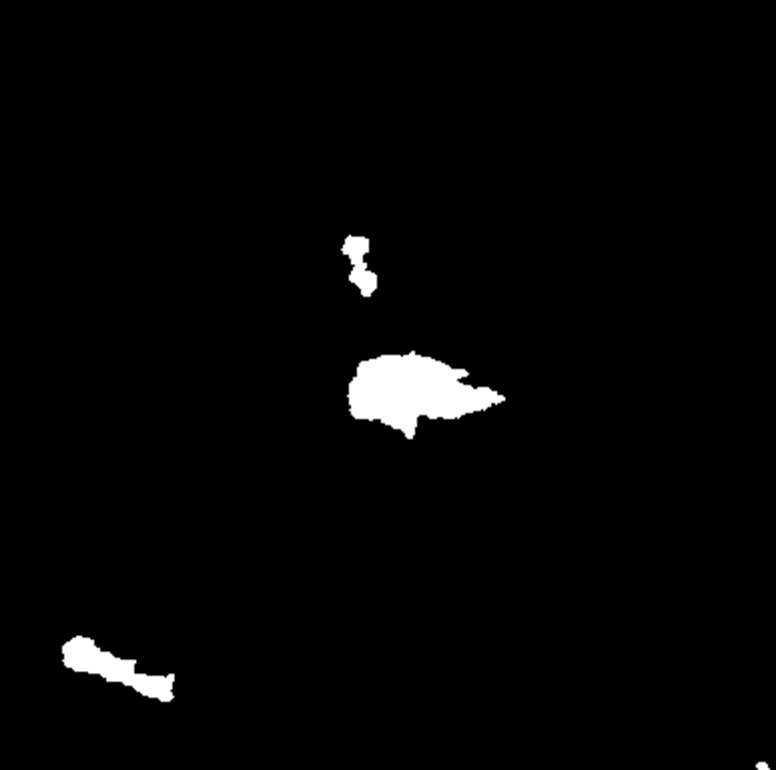}}
    
    \subfloat[\gls{sar}  Image 2]{\includegraphics[width=0.4\textwidth]{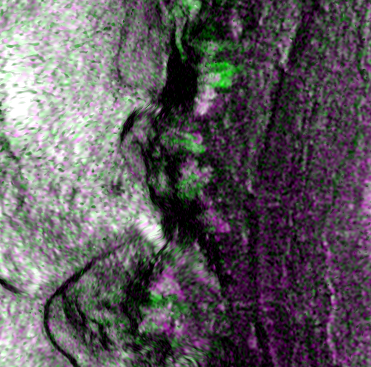}}
    \hspace{0.1cm}
    \subfloat[Ground Truth 2]{\includegraphics[width=0.4\textwidth]{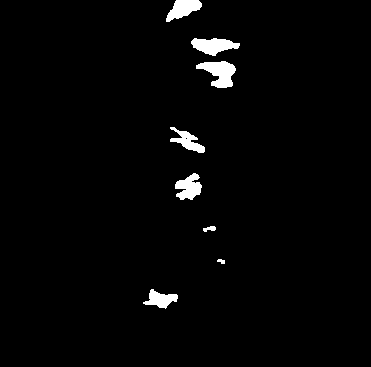}}
    
    \caption{Avalanche segmentation: (\textbf{a,c}) \gls{sar} backscatter images created through Algorithm \ref{alg:sar_rgb} discussed in Appendix \ref{manual_segmentation} and (\textbf{b,d}) corresponding ground truth masks.}
    \label{fig:sar_comparison}
\end{figure}

Mapping avalanche activities is a crucial component to aid forecasting and risk mitigation in mountainous regions~\cite{eckerstorfer2017complete}.
Every year, more than $100$ avalanche-related fatalities are reported across Europe, and numerous infrastructures, roads, and buildings are damaged by this phenomenon~\cite{paper2}.
On-field measurements are the most reliable option to quantitatively assess the area covered by an avalanche, but these are expensive, limited by accessibility, risky for observers, and unsuitable for broad and continuous monitoring~\cite{eckerstorfer2019near}. 
Remote sensing has therefore emerged as a valuable alternative, enabling safe, large-scale, and frequent monitoring of avalanche activity through the systematic acquisition of high-quality satellite imagery, particularly at high latitudes \cite{eckerstorfer2016remote}.

Among the available remote sensing modalities, \acrfull{sar} data is especially well suited for avalanche detection, as it is independent from weather conditions and exhibits clear scattering patterns of snow debris~\cite{grahn2024data}.
However, the manual identification of snow avalanches in \gls{sar} images is complex, time-consuming, and requires expert knowledge~\cite{eckerstorfer2015manual}.
To overcome these limitations, several automated approaches for continuous avalanche monitoring have been proposed in recent years~\cite{vickers2016method}.
In particular, deep learning-based image segmentation methods currently represent the most promising solutions. Nevertheless, existing models still suffer from a high rate of false positives and fail to reach the same level of accuracy as human experts \cite{bianchi2026snow}. 
Therefore, manual annotation of \gls{sar} images still represents the gold standard in the field~\cite{eckerstorfer2016remote}. 

Perhaps the major obstacles limiting further improvements of deep learning methods is the scarcity of labeled data, which is necessary to train more accurate models. 
Manual annotations are not only costly to produce, but also prone to errors: smaller avalanches are often overlooked, while imprecise drawing of the mask contours introduces label noise that negatively influences the performance of segmentation models~\cite{paper, paper2}. 
These inaccuracies arise from a combination of annotator subjectivity, speckle noise in the \gls{sar} images, and ambiguity in interpreting the actual contours of the debris.

The goal of this study is to develop a tool that facilitates the annotation of snow avalanches and improves the quality of the segmentation masks in \gls{sar} imagery.
To this end, we adapt \gls{sam}~\cite{paper1} to the task of avalanche annotation and evaluate its integration into a semi-automatic annotation workflow. 
\gls{sam} is a computer vision foundation model, able to identify and segment any object in natural images with remarkable accuracy, requiring minimal user inputs in the form of prompts: simple clicks or \glspl{bb} drawn around the objects of interest.
Rather than relying on explicit class information, \gls{sam} leverages prompt-based guidance to localize the target object, which makes it highly flexible and transferable to downstream tasks with minimal retraining.
However, images from the \gls{sar} domain significantly differ from RGB images on which \gls{sam} was originally trained, preventing its straightforward application for avalanche segmentation in \gls{sar} images.
Recent studies have demonstrated that adapting \gls{sam} to specialized imaging modalities can substantially reduce annotation effort while maintaining high segmentation quality~\cite{paper17, paper16, paper18, paper20}. 
These findings motivate the exploration of \gls{sam} as a semi-automated annotation tool for snow avalanches in \gls{sar} data.

The main contribution of our work is to extend \gls{sam} to our use-case by addressing the following key challenges:

\begin{itemize}
    \item \textbf{Domain adaptation:} Given the limited amount of training data, effective domain adaptation must be achieved by fine-tuning only a small subset of \gls{sam}'s parameters.
    \item \textbf{Input adaptation:} The standard \gls{sam} architecture can only process three-channels inputs, whereas raw \gls{sar} images consist of a different number of channels.
    \item \textbf{Improved prompt robustness:} \gls{sam} struggles with small targets and with imprecise prompts, e.g., \glspl{bb} much larger than the object of interest. Identifying prompt strategies that improve robustness, especially for segmenting small avalanches, is therefore critical.
    \item \textbf{Training optimization:} Given that even the lightest \gls{sam} variant exceeds $90$ million parameters, the fine-tuning procedure must be carefully designed to be computationally feasible on commercial hardware.
\end{itemize}

To address these challenges, we \emph{i}) employ Adapters~\cite{pu2025classwise} in combination with decoder fine-tuning for tackling domain shifts, \emph{ii}) introduce a multiple encoder method inspired by~\cite{xiao2024segment} to process the six channels of \gls{sar} images, \emph{iii}) propose a specific prompt strategy based on \glspl{bb}, and \emph{iv}) introduce a custom algorithm for training \gls{sam}.

Experimental results demonstrate that our approach successfully adapts \gls{sam} to the \gls{sar} domain, achieving competitive or superior performance compared to existing methods in the literature.
Moreover, the proposed prompt strategy reduces sensitivity to prompt precision, enabling performance comparable to prompt-free segmentation approaches when using a full-image (minimum precision) prompt.
Finally, the integration of our method into a semi-automatic annotation tool significantly improves annotation efficiency, demonstrating its practical value for generating a large-scale inventory of snow avalanches.

\section{Related Work}

In this section, we provide a few essential notions on \gls{sar} remote sensing (Section \ref{subsec:rw_sar_sensing}), we mention the most effective solutions for segmenting avalanches in \gls{sar} images (Section \ref{subsec:rw_avalance_segmentation}), and then we introduce \gls{sam} and \gls{sam} adaptation (Section \ref{subsec:rw_sam} and \ref{subsec:rw_sam_adaptation}, respectively).

\subsection{\acrlong{sar} Images for Avalanche Mapping}
\label{subsec:rw_sar_sensing}

\gls{sar} images are obtained from the backscattered energy of the microwave signals emitted by the radar itself.
Unlike optical sensors that are passive and capture reflected sunlight, \gls{sar} is an active technology that operates independently of sunlight or cloud cover. 
At typical operating frequencies (e.g., X-, C-, or L-band), \gls{sar} sensors are highly sensitive to surface properties, such as roughness and moisture. 
Therefore, debris deposited by snow avalanches can be distinguished from the surrounding undisturbed snow because of their different roughness and structural properties, resulting in an increased, enabling detection in \gls{sar} images~\cite{paper2,paper}. 

\gls{sar} sensors operate using different polarization modes, which describe the orientation of the transmitted and received electromagnetic waves. 
These are generally categorized into co-polarized signals (\texttt{VV} or \texttt{HH}), where the transmit and receive orientations are the same, and cross-polarized signals (\texttt{VH} or \texttt{HV}), where they are orthogonal. 
While some ``quad-polarized'' satellites can acquire all four combinations, Sentinel-1 operates in a dual-polarized mode. 

In the context of snow avalanche observation, the signal is primarily composed of:

\begin{itemize}
    \item \texttt{VV} (vertical transmit, vertical receive): which is the most sensitive to rough surface scattering and the most informative data source for avalanche segmentation~\cite{paper}.
    \item \texttt{VH} (vertical transmit, horizontal receive): which is the most sensitive to volume scattering and often used to complement the \texttt{VV} channel.
\end{itemize}

Human experts annotate avalanche debris by looking at RGB composites obtained by combining two co-registered \gls{sar} images taken at consecutive times $t_0$ and $t_1$.
The time offset between two different passes can vary between $6$ and $12$ days.
The most common RGB composites are given by $[\texttt{VV}_0, \texttt{VV}_1, \texttt{VV}_0]$ or created through specific algorithms like those described in Appendix~\ref{manual_segmentation}.
Unfortunately, \gls{sar} images are affected by speckle noise, which negatively influences their use for many tasks. 
Speckle noise can be reduced during image preprocessing, e.g., by applying the Lee filter~\cite{paper46}. 
Modern noise-removal approaches consist of pretraining deep learning models in a self-supervised way to reduce the impact of speckle noise on the final performance~\cite{paper22, paper24}. 

Useful auxiliary inputs that aid the human annotator to perform manual segmentation are the \gls{dem} and the \gls{met} data. 
\gls{dem} images associate with each pixel a real value representing the elevation above the sea level expressed in meters.
In the context of snow avalanche detection, \gls{dem} data can inform the model about areas where avalanches cannot occur, i.e., flat surfaces far from mountain slopes. 
The \gls{dem} can be used to derive the \gls{sa}, a topographical feature that can be used to identify release zones, i.e., those regions where avalanches can release debris.
In particular, avalanche debris can be found in the proximity of slopes whose inclination ranges between $30-50$ degrees~\cite{paper}.
The \gls{sa} is defined as:
\begin{equation}
    \label{eq:Slope Angle}
    \theta = \arctan(|\nabla E(p)|),
\end{equation}
where $E(p)$ is the elevation associated with pixel $p$ and $\nabla E(p)$ represents the gradient.

The relevance of \gls{met} data for avalanche detection has been highlighted in previous work~\cite{paper5,paper3,paper2}, but its impact on performing automatic segmentation is still to be determined.
The \gls{met} data consists of a time series associated with each \gls{sar} image, which spans the entire duration $[t_0, t_1]$ between the two satellite acquisitions.

\subsection{Automated avalanches detection with Deep Learning}
\label{subsec:rw_avalance_segmentation}

Automatic detection of snow avalanches with deep learning is still a relatively new field.
Deep learning approaches must reach a certain degree of reliability before being deployed in avalanche warning services to assess the avalanche danger and support decision-making in specific communities~\cite{paper2}.
The most prominent deep learning model for segmentation by Bianchi \emph{et al.}~\cite{paper} is based on a U-net with an encoder-decoder structure and skip connections, which takes as input \gls{sar} images, the terrain slope, and other topographic features. 
While the model achieved performance superior to existing approaches in automated avalanche detection, it still produces several detections not corresponding to annotations in the test set.
Despite most of them being false alarms, some of the false positives were actual avalanches missed during the labeling by the expert, highlighting a limitation in the manual annotation process.

\subsection{\acrlong{sam}}
\label{subsec:rw_sam}

\gls{sam} \cite{paper1} is a computer vision foundation model that can identify and segment almost any object in natural RGB images, achieving remarkable accuracy.
\gls{sam} does not leverage any class information but, instead, relies on a minimal user prompt to isolate objects from their background, thus generating binary masks. 
This makes \gls{sam} adaptable to many downstream tasks with minimal to no retraining, often through prompt engineering (given the right prompt, the model generalizes well even on unseen objects and different domains). 
Prompts can be points, \glspl{bb}, masks, and text. 
\glspl{bb} in particular are represented with a tuple $[x_1, y_1, x_2, y_2]$ that corresponds to the top-left and bottom-right corners.

The original \gls{sam} was trained using a data engine technique composed of three subsequent phases: assisted manual, semi-automatic, and fully-automatic.
The procedure progressively reduces the presence of a human in the loop, leading to the creation of the SA-1B dataset~\cite{paper1}, the largest dataset currently available for segmentation on natural images.

\begin{figure}
    \centering
    \includegraphics[width=.9\linewidth]{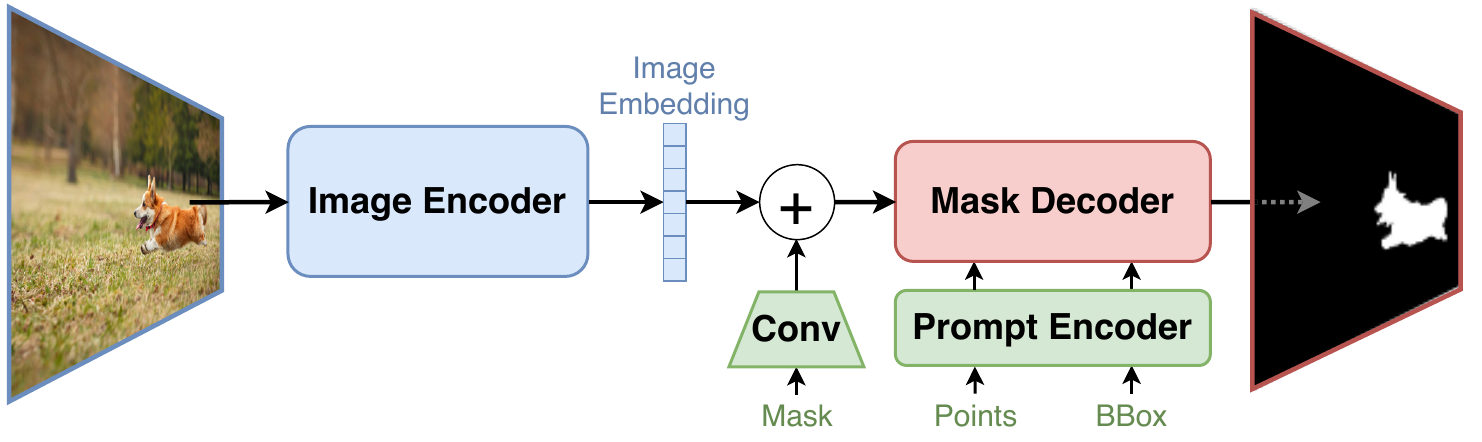}
    \caption{Overview of the \gls{sam} architecture. 
    A heavyweight image encoder outputs an image embedding, given an RGB image. 
    The prompt encoder identifies the segmentation target, which is then segmented by the mask decoder.}
    \label{fig:sam}
\end{figure}

The architecture of \gls{sam}, depicted in Figure~\ref{fig:sam}, is composed of:

\begin{itemize}
    \item \textbf{Image Encoder}. 
    A classical pretrained \gls{vit}~\cite{dosovitskiy2021an} which takes as input $1024 \times 1024$ RGB images and outputs $256\times64\times64$ embeddings.
    \item \textbf{Prompt Encoder}. 
    Prompts can be either sparse (points, boxes, and text) or dense (segmentation masks). Sparse prompts are mapped to $256$-dimensional embeddings and used later for computing cross-attention with the image embedding in the decoder. 
    Dense prompts are processed with convolutional layers and directly mapped to the image embedding.
    \item \textbf{Mask Decoder}. 
    The decoder takes as input the image embedding and the encoded sparse prompts and outputs a probability map. The decoder architecture updates both the image and prompt embeddings by relying on self-attention applied to the prompt embedding and on bidirectional cross-attention between the image features and the prompt embeddings.
\end{itemize}

There are three main versions of \gls{sam}, which differ in the number of parameters:
\begin{itemize}
    \item \gls{vit}-B: $91$ million parameters.
    \item \gls{vit}-L: $308$ million parameters.
    \item \gls{vit}-H: $636$ million parameters.
\end{itemize}
The choice between these versions is driven by the target latency, hardware requirements, and the training set size in case of \gls{sam} re-training.

\subsection{Adapting \gls{sam}}
\label{subsec:rw_sam_adaptation}

The literature presents several solutions for adapting \gls{sam} to different domains. 
Med-\gls{sam} consists of a successful retraining of \gls{sam} to medical images, resulting in an effective tool for assisting doctors and medical experts~\cite{paper17}. 
In Med-\gls{sam}, both the \gls{sam} encoder and the decoder were finetuned on a very large amount of labeled medical images. 
This re-training strategy is not viable in our specific case due to the shortage of annotated images and computational constraints. 

Another successful approach to adapt \gls{sam} consists of substituting the decoder. 
This strategy allowed applying \gls{sam} to the \gls{sar} domain and enabled the construction of SAMRS, the largest dataset for semantic segmentation in remote sensing~\cite{paper18}.
Substituting the decoder, usually paired with encoder adaptation, enables multi-class segmentation but gives up the possibility of using prompts~\cite{paper13, paper20, paper18, pu2025classwise}. 
It is possible to modify the decoder of \gls{sam} to perform multi-class segmentation while still allowing prompts by changing the convolutional layers of the decoder. 
However, this modification requires fundamental changes also to the prompt handling, and to our knowledge, it has not been applied in practice. 
In our study, we did not explore substitution or major modifications of the decoder for two reasons:
\begin{itemize}
    \item Avalanche segmentation can be cast as a binary segmentation problem, where avalanches play the role of the foreground class.
    \item We wanted to preserve prompts, which are fundamental for semi-automatic segmentation.
\end{itemize}
Since the \gls{sam} image encoder is characterized by a large number of parameters and layers, it represents a significant computational bottleneck. 
As a consequence, full fine-tuning of \gls{sam} is often an impractical solution for domain adaptation. 

Recent literature focuses on \gls{peft} strategies to adapt \gls{sam} to a new domain. 
Many methods have been proposed in this direction, and the most popular ones are \gls{lora}~\cite{paper9}, Adapters~\cite{pu2025classwise}, and Auto-\gls{sam}~\cite{paper11}.
The latter tries to adapt \gls{sam} to a new domain through the introduction of a parallel network and belongs to another family of methods that tries to improve \gls{sam} performance on new domains by adding additional prompts or modifying existing ones~\cite{paper20, paper11, xiao2024segment, paper16}.

\section{Materials and Methods}

Section~\ref{sec:dataset} describes the avalanche dataset, the available modalities (multi-temporal \gls{sar} channels, \gls{dem}, and derived \gls{sa}), and the preprocessing needed to meet \gls{sam}'s fixed input resolution.
Section~\ref{sec:domain_adapt} details how we adapt \gls{sam} to the \gls{sar} domain by training lightweight Adapters in the \gls{vit} image encoder and fine-tuning the mask decoder.
Section~\ref{sec:resource_opt} presents a compute-efficient training scheme that reuses image embeddings for all prompts associated with the same image.
Section~\ref{sec:prompt_robustness} describes our \gls{bb}-based prompting strategy, including prompt generation from masks and augmentation to handle imprecise user inputs.
Section~\ref{sec:input_adaptation} introduces our multi-encoder architecture to leverage a larger number of input channels, based on supervised embedding alignment and fusion.
Section~\ref{sec:training_method} then combines these elements into the final three-phase training procedure and reports the main optimization settings.
Finally, Section~\ref{sec:segmentation:tool} presents the web-based tool used to assess the method in a human-in-the-loop annotation workflow. 

\subsection{Dataset}
\label{sec:dataset}

The dataset consists of $2,681$ labeled samples acquired over various regions in Norway. 
Each sample maintains a ground sampling distance of $10\text{ m} \times 10\text{ m}$ per pixel, with image resolutions ranging from $355 \times 363$ to $512 \times 512$ pixels. 
Each observation comprises three distinct data modalities: \gls{sar}, \gls{dem}, and \gls{met}. 
Our preliminary results showed that prompting \gls{met} data to the segmentation model did not improve the performance, despite the relevance of this data in avalanche detection, see Appendix~\ref{meteo_data}. 
Therefore, \gls{met} data are not further discussed in the following.

The \gls{sar} data is represented by the two \gls{sar} images, with both the \texttt{VV} and \texttt{VH} channels, collected at time steps $t_0$ and $t_1$.
In our dataset, the two images are taken either at $6$ or $12$ days apart.
To create an RGB image composite, each \gls{sar} channel must also be rescaled from the original $dB$ scale to $[0, 1]$.
For these specific datasets, the images used for manual labeling were created through Algorithm \ref{alg:sar_rgb}. 
We defer the algorithm and the details on manual labeling to Appendix \ref{manual_segmentation}.

The values of the single channel \gls{dem} images in our dataset range from $19.11$ meters to $2274.41$ meters with a mean value of $675.82$ meters and a standard deviation of $380.62$ meters.
The images must be rescaled to make them compatible with RGB standard values ($[0,255]$ integer or $[0,1]$ float).
We also processed \gls{dem} images to derive the \gls{sa} values expressed in angular degrees.

To satisfy the fixed input requirement of the \gls{sam} image encoder ($1024 \times 1024$ pixels), each sample was resized such that its longest dimension matched the target resolution. 
For samples with smaller aspect ratios, the remaining area was zero-padded, ensuring the preservation of the original spatial proportions and preventing geometric distortion of the \gls{sar} and \gls{dem} features.

\subsection{Domain adaptation}
\label{sec:domain_adapt}

As previously discussed, \gls{sam} was originally trained on RGB images from the natural image domain, which substantially differs from the \gls{sar} and \gls{dem} data in our dataset.
Adapting \gls{sam} to avalanche segmentation requires a fine-tuning step in which a subset of the model parameters is retrained. 
Training the entire model would lead to severe risks of over-fitting, given the limited size of our dataset and the large number of model parameters (the smallest version of \gls{sam} used in our experiments is based on \gls{vit}-B and contains over $91M$ parameters). 
On top of that, fine-tuning \gls{sam} requires a great computational effort.
In the following, we separately discuss how we adapted the encoder and decoder components of \gls{sam}.

\subsubsection{Image encoder}
In \gls{sam}, most of the parameters are concentrated in the image encoder, which in the model based on \gls{vit}-B, comprises approximately $86M$ parameters and represents the major bottleneck to perform both training and inference.
As discussed in Section~\ref{subsec:rw_sam_adaptation}, several methods have been proposed in the literature to adapt \gls{sam}'s image encoder and, in general, large foundation models to different domains. 
In this work, we experimented with Auto-\gls{sam}~\cite{paper11}, \gls{lora}~\cite{paper9}, and Adapters~\cite{pu2025classwise}.
Among these, we found that the Adapters yielded the best performance in terms of \gls{iou} metric on the avalanche class, which served as our main validation metric.

Adapters are trainable components that are placed in the transformer block of the \gls{vit} between the multi-head attention and the residual connection, and in parallel with the \gls{mlp} layer, as shown in Figure~\ref{fig:Adapters}. 
These modules transform the intermediate hidden states $x$ of the \gls{vit} as follows:
\begin{equation}
    \text{Adapter}(x) = \texttt{Up}(\texttt{ReLU}(\texttt{Down}(x)))
    \label{eq:adapter}
\end{equation}
where \texttt{ReLU} is the standard activation function, and \texttt{Up}$(\cdot)$ and \texttt{Down}$(\cdot)$ represent two fully connected layers that perform upscaling and downscaling, respectively.
\begin{figure}
    \centering
    \includegraphics[width=0.45\linewidth]{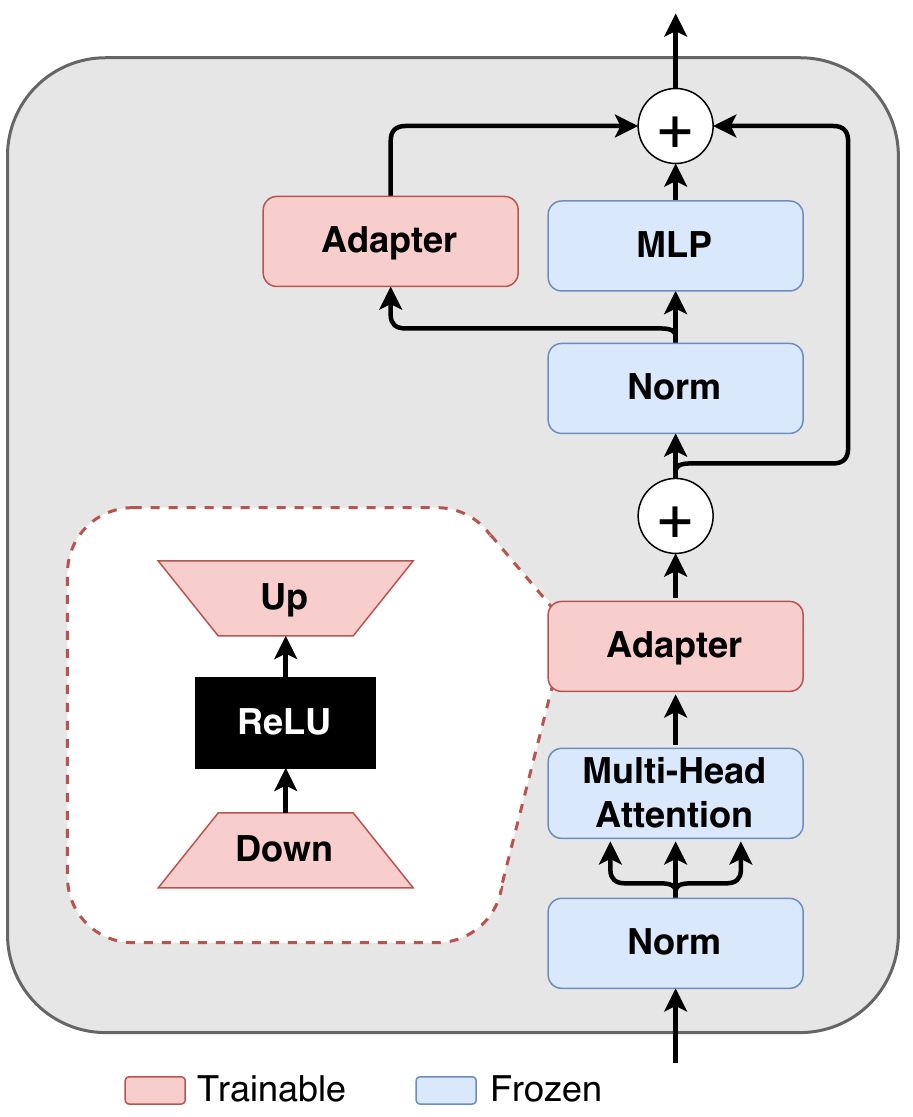}
    \caption{Transformer block of the \gls{vit} modified with adapters, composed of two linear layers and an activation function, positioned after the multi-head attention and in parallel with the \gls{mlp}.}
    \label{fig:Adapters}
\end{figure}

Adding Adapters to every Transformer block of the \gls{vit}-B introduces $7M$ parameters in the \gls{vit}-B encoder, which consists of only $10\%$ of the encoder parameters, reducing by over $90\%$ the number of trainable parameters with respect to fine-tuning. 
We did not find a benefit in using dropout and set the \gls{mlp}-ratio, which is the ratio between the number of output and input neurons of the linear layer, to $0.25$.

Following the standard implementation in the PyTorch \texttt{Linear} layer, the weights and biases in the \texttt{Up} and \texttt{Down} layers of the Adapters are initialized from a uniform distribution $U(-k, k)$, where the parameter $k$ is defined as:
\begin{equation*}
    k = \frac{1}{\sqrt{\text{in\_features}}}\,.
\end{equation*}
This is a rather generic and uninformative weight initialization, which usually works best when there are many training samples available to learn the best weight configuration.
However, preliminary results showed that initialization schemes more tailored to our use case, including pre-training the adapters with self-supervised objectives, did not convey significant improvements in our experiments. 
Additional details on pre-training and weight initialization are discussed in Appendix~\ref{self-supervised_pretraining}.

\subsubsection{Decoder}
Since the decoder already outputs a binary mask, which in our case serves as the avalanche class, we did not have to change the architecture of the decoder. Therefore, we simply fine-tuned the decoder as done in previous work~\cite{pu2025classwise, paper18, paper20}. 
Plain fine-tuning without modifying the decoder architecture allows us to preserve the prompt for the semi-automatic annotation, which is the main use case for our model.

\subsection{Robustness to inaccurate prompts}
\label{sec:prompt_robustness}

\begin{figure}
    \centering
    \includegraphics[width=1\linewidth]{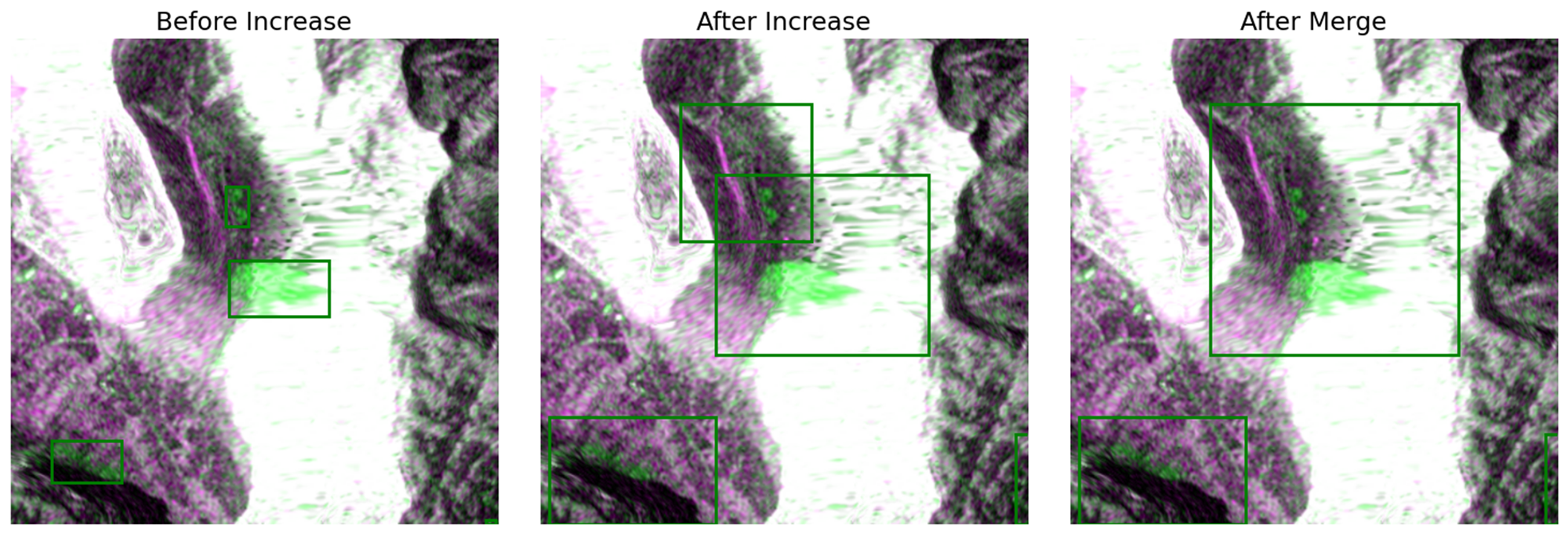}
    \caption{Creation of \glspl{bb} to improve robustness to inaccurate prompts. 
    From left to right, we shown: the creation of the minimum enclosing \gls{bb} from the segmentation mask, the random increase of the \gls{bb} dimensions, and the merging of overlapping \glspl{bb}.}
    \label{fig:bbcreation}
\end{figure}

We adopt \glspl{bb} as prompts for \gls{sam}, as they are intuitive, easy to provide, and widely recognized as the most effective prompting strategy for semi-automatic annotation, particularly in the context of \gls{sar} imagery~\cite{paper18}.
We created \glspl{bb} from the segmentation masks by: 
(i) Computing for each avalanche the minimum enclosing rectangle;
(ii) Increasing the \glspl{bb} through an ad-hoc augmentation strategy;
(iii) Merging \glspl{bb} that intersect to increase efficiency and simulate a more realistic human input. 
All three steps are illustrated in Figure \ref{fig:bbcreation}.

Since \gls{sam} does not explicitly leverage any class information, the prompt alone determines the target object to be segmented.
In the context of avalanche segmentation, we noticed that inaccurate \glspl{bb} lead to a drop in segmentation performance, suggesting that inaccurate localization hampers the model's ability to correctly identify avalanche debris.
In an operational setting, however, some degree of imprecision in human-provided prompts is unavoidable.
To improve \gls{sam} robustness when inaccurate \glspl{bb} are prompted, our prompt strategy increases the \gls{bb} by displacing the four coordinates with a random value drawn from a uniform distribution $U(0, k)$, where $k$ denotes the maximum number of offset pixels.
During training, we used a mixed prompt strategy in which $80\%$ of the prompts were \emph{accurate} boxes ($k = 40$), $10\%$ were \emph{inaccurate} ($k = 200$), and in $10\%$ of cases the \glspl{bb} were replaced by \emph{full-image} \glspl{bb}. 
To perform model selection in the validation stage, we only used accurate boxes: instead of drawing the displacement from the uniform distribution $U(0, 40)$, we use a fixed value of $20$. 
We observed that the introduction of the inaccurate and full-image \glspl{bb} during training made the model more robust to imprecise prompts, without affecting the usage of accurate prompts in validation.

On top of improving robustness to inaccurate prompts, we found that our strategy yields two additional benefits.
The first is that training with imprecise and full-image prompts improves the segmentation performance over small avalanches. 
This is a particularly important result since the predictions from the baseline model exhibit a positive correlation between avalanche size and the \gls{iou}, indicating that smaller avalanches are more challenging to segment.
In Appendix \ref{small_optimization}, we discuss additional details and attempts for improving the detection performance on small avalanches. 
The second benefit is that the proposed augmentation strategy trains the model to perform full-image segmentation, enabling prompt-free inference.

\subsection{Resources optimization}
\label{sec:resource_opt}

As discussed in Section \ref{sec:domain_adapt}, we employ the \gls{vit}-B variant of \gls{sam}, which contains $91$ million parameters, $86$ million of which reside in the image encoder and constitute the primary computational bottleneck during fine-tuning. 
In \gls{sam}, the image encoder and the prompt encoder operate independently: the image embedding depends only on the input image, while the prompt embedding depends exclusively on the provided prompt (e.g., the \gls{bb}). 
This architectural property becomes particularly relevant in our setting, where multiple \gls{bb} prompts are associated with different avalanches appearing in the same image.
We note that this differs from the more common setting where a natural image contains a single object of interest.

Processing each image-prompt pair independently would result in redundant and computationally expensive recomputation of the same image embedding.
To avoid this redundancy, we compute the image embedding only once per image and reuse it for all associated prompts. 
Algorithm~\ref{alg:train_alg} details the data preparation procedure that enables the resource optimization. 
The key idea is to replicate the image embedding (Line~\ref{line:repeat} of Algorithm~\ref{alg:train_alg}) so that it can be paired with each prompt and processed in parallel by the decoder. 
This allows all prompts associated with the same image to be evaluated simultaneously, significantly improving training efficiency. 
All the repeated image embeddings are then concatenated to form a single expanded batch (Line~\ref{line:concat1}), which is fed to the decoder together with the corresponding concatenated prompt embeddings (Line~\ref{line:concat2}).

\begin{algorithm}[H]
\caption{Data preparation for resources optimization.}
\label{alg:train_alg}
\begin{algorithmic}[1]
\Input
\emph{Image embeddings} $\mathcal{I} = \{z_i\}_{i=1}^{B}$, where $z_i \in \mathbb{R}^{C \times H \times W}$ is the embedding of the $i$-th image and $B$ is the batch size, \emph{Prompt embeddings} $\mathcal{P} = \{p_i\}_{i=1}^{B}$, where $p_i \in \mathbb{R}^{L_i \times 2 \times C}$ and $L_i$ is the number of prompts for the $i$-th image
\Output Expanded image embeddings $\hat{\mathcal{E}}$ and concatenated prompt embeddings $\hat{\mathcal{P}}$
\Function{PrepareDecoderInput}{$\mathcal{I}, \mathcal{P}$}
    \State $K \leftarrow \sum_{i=1}^B{L_i}$ \hfill \Comment{Total number of prompts}
    \State $\hat{\mathcal{I}} \leftarrow []$
    \State $\hat{\mathcal{P}} \leftarrow []$
    \For{$i = 1, \dots, B$}
        \State $\hat{z}_i \leftarrow \Call{Repeat}{z_i, L_i}$ \hfill \Comment{New shape: $L_i \times C \times H \times W$} \label{line:repeat}
        \State $\hat{\mathcal{I}} \leftarrow \Call{Concat}{\hat{\mathcal{I}}, \hat{z}_i}$ \label{line:concat1}
        \State $\hat{\mathcal{P}} \leftarrow \Call{Concat}{\hat{\mathcal{P}}, p_i}$ \label{line:concat2}
    \EndFor
    \State \Return $\hat{\mathcal{I}}, \hat{\mathcal{P}}$ \hfill \Comment{Final shapes: $K \times C \times H \times W$ and $K \times 2 \times C$}
\EndFunction
\end{algorithmic}
\end{algorithm}

Depending on the number of prompts, the proposed parallelization could occupy a lot of memory, but even if each image must be processed individually, the overall compute time is reduced.
Indeed, handling prompts at run time efficiently allowed us to train simultaneously on all the avalanches of the same image and reduced the training time by approximately $63\%$ without impacting the number of epochs needed to reach convergence in training.

\subsection{Input adaptation}
\label{sec:input_adaptation}

The \gls{sam} image encoder was pre-trained on RGB images and expects a three-channel input.
Our avalanche dataset, however, provides six co-registered channels: $\texttt{VV}_0$, $\texttt{VV}_1$, $\texttt{VH}_0$, $\texttt{VH}_1$, \gls{dem}, and \gls{sa}.
To exploit all available information without altering the pre-trained encoder architecture, we adopt a multi-encoder strategy inspired by \gls{samm}~\cite{xiao2024segment}.

Concretely, we use two \gls{sam} image encoders that process complementary triplets: a \emph{primary} encoder fed with $[\texttt{VV}_0, \texttt{VV}_1, \gls{dem}]$ and a \emph{secondary} encoder fed with $[\texttt{VH}_0, \texttt{VH}_1, \gls{sa}]$.
Both encoders share the same backbone architecture and are adapted with the same \gls{peft} mechanism described in Section~\ref{sec:domain_adapt}.
For a batch of size $B$, the two image encoders produce embeddings $e_1, e_2 \in \mathbb{R}^{B \times 256 \times 64 \times 64}$.

A key requirement of this design is that the embeddings produced by the two encoders are compatible with a single mask decoder, so that they can be fused and decoded consistently.
In the following, we describe our task-aware alignment strategy and the fusion mechanism used to combine the aligned embeddings.

\subsubsection{Embeddings Alignment}
\label{embed_alignment}

In \gls{samm}, the auxiliary encoder is aligned to a frozen primary encoder by minimizing a distance metric between their embeddings (e.g., \gls{mse}), an unsupervised objective often referred to as \emph{embedding unification}.
While this facilitates combining modalities, it also encourages the secondary encoder to reproduce information already present in the primary representation.
This is not necessarily optimal for segmentation, where the goal is to extract complementary features that improve the final mask prediction.

We instead align the secondary encoder to the task-specific representation learned by the primary model.
After adapting the primary model to \gls{sar} avalanche segmentation (Section~\ref{sec:domain_adapt}), we freeze its mask decoder and train the secondary encoder using the supervised segmentation loss computed on the decoder output.
Because the decoder parameters are fixed, the secondary encoder must generate embeddings that lie in the same space expected by the decoder, enabling subsequent fusion while preserving complementary information from the secondary modality.
In our experiments, this supervised alignment strategy yielded the best performance among the considered input adaptation variants (Appendix~\ref{app:input_adaptation}).

\subsubsection{Embedding Fusion}
\label{sec:embedding_fusion}

Once both encoders produce aligned embeddings, we fuse them at the embedding level.
As a simple baseline, we use a global convex combination:
\begin{equation}
    \label{eq:sfg2}
    \hat{e}_F = \alpha \cdot e_1 + (1-\alpha) \cdot e_2
\end{equation}
With $\alpha = 0.5$, this baseline already improves over training on a single modality, confirming that the supervised alignment enables complementary information to be exploited.

To allow the relative contribution of each modality to vary spatially, we introduce a \gls{sfg} (Figure~\ref{fig:sfg}).
The \gls{sfg} predicts an element-wise weight tensor $\omega \in [0, 1]^{B \times 256 \times 64 \times 64}$ from the concatenation of $e_1$ and $e_2$ and computes the fused embedding as:
\begin{equation}
    \label{eq:sfg}
    \hat{e}_F = \omega \odot e_1 + (1-\omega) \odot e_2,
\end{equation}
where $\odot$ denotes element-wise multiplication.

\begin{figure}[H]
    \centering
    \includegraphics[width=0.6\linewidth]{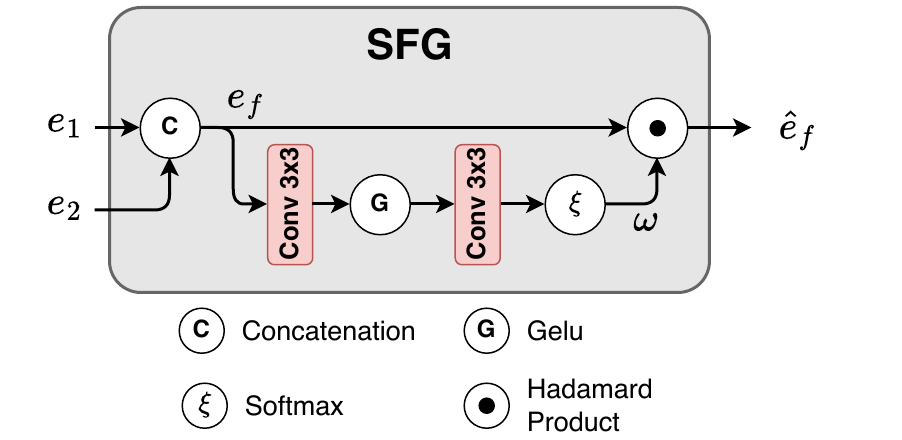}
    \caption{\acrlong{sfg}: given two image embeddings $e_1$ and $e_2$, the gate predicts weights $\omega$ from their concatenation and produces the fused embedding $\hat{e}_F$.}
    \label{fig:sfg}
\end{figure}

We note that other input adaptation strategies, including channel selection and patch-embedding modifications, were also investigated but did not yield comparable performance improvements. 
Further details can be found in Appendix~\ref{app:input_adaptation}.

\subsection{Training Procedure}
\label{sec:training_method}

\begin{figure}
    \centering
    \includegraphics[width=0.7\linewidth]{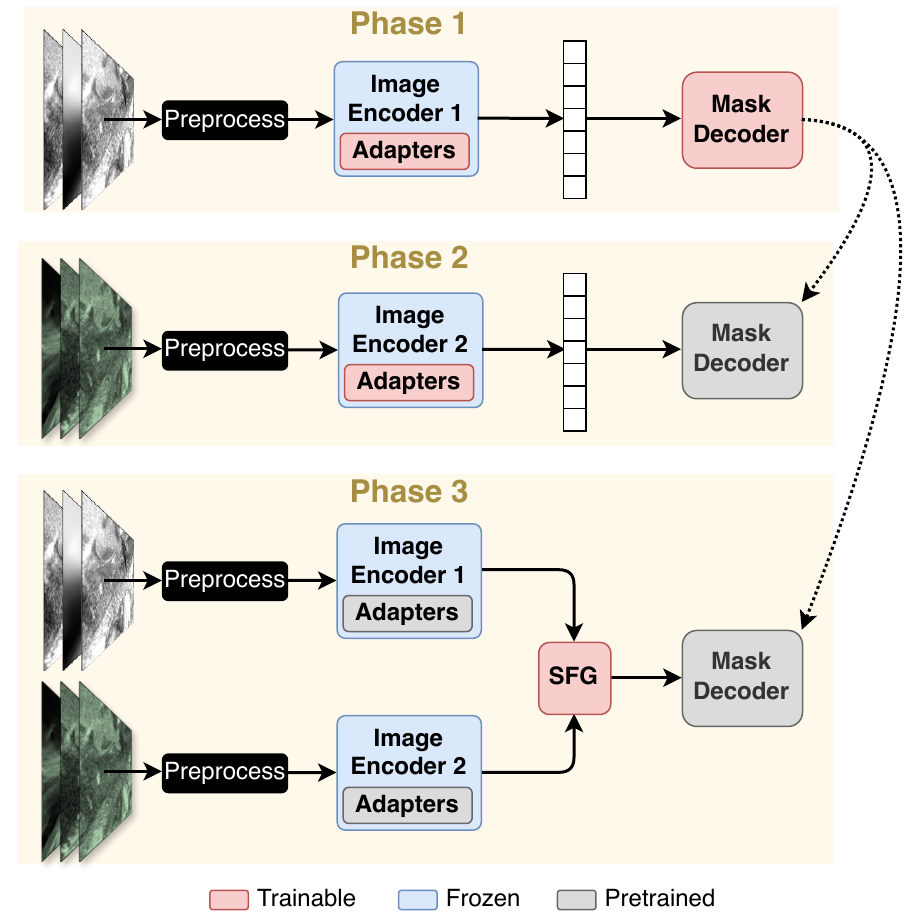}
    \caption{Overview of the training procedure: in the first phase, a model with $\texttt{VV}_0$, $\texttt{VV}_1$ and \gls{dem} as input is trained with adapters and its decoder fine-tuned, using our prompt and efficient parallelization strategies. 
    In the second phase, a model with $\texttt{VH}_0$, $\texttt{VH}_1$, and \gls{sa} as input is trained using the supervised embedding alignment strategy. 
    The third phase combines the image embeddings through a \gls{sfg}.}
    \label{fig:finalmodel}
\end{figure}

The overall training procedure consists of three sequential phases, illustrated in Figure~\ref{fig:finalmodel} and detailed below.

\begin{itemize}
    \item \textbf{Phase 1 - Primary modality adaptation.} 
    The primary model is trained on $[\texttt{VV}_0, \texttt{VV}_1, \gls{dem}]$.
    As discussed in Section~\ref{subsec:rw_sar_sensing}, the \texttt{VV} polarization is the most informative \gls{sar} source for avalanche mapping and is used to create RGB composites for manual annotation; we therefore use $[\texttt{VV}_0, \texttt{VV}_1]$ as the primary \gls{sar} inputs and complement them with \gls{dem}.
    The model used at this stage leverages the approaches discussed in Section~\ref{sec:domain_adapt} (adapters-based encoder tuning and decoder fine-tuning), Section~\ref{sec:prompt_robustness} (prompt-robust training), and Section \ref{sec:resource_opt} (resource optimization). 
    This supervised training stage is necessary to account for the domain shift arising from natural RGB images to \gls{sar}.
    
    \item \textbf{Phase 2 - Secondary modality alignment.} 
    A secondary model is trained to extract image embeddings from the remaining three input channels ([$\texttt{VH}_0$, $\texttt{VH}_1$, \gls{sa}]), in a supervised manner.
    As described in Section~\ref{embed_alignment}, we force alignment to the same embedding space through the frozen decoder of the primary model trained in Phase 1 (Figure \ref{fig:finalmodel}).
    Freezing the decoder reduces the individual performance of the secondary model, since we only train the adapters, but facilitates the combination of the embeddings later on, which is the main goal. 
    Moreover, this secondary modality is supposed to complement the main modality, which justifies using the decoder from the primary modality. 
    
    \item \textbf{Phase 3 - Embedding fusion.} 
    In the final phase, a \gls{sfg} is trained to combine the embeddings produced by the two encoders. 
    Once again, we perform supervised training, and we leverage the frozen decoder of the primary model trained in Phase 1 (Figure \ref{fig:finalmodel}).
\end{itemize}

Experiments were carried out using \texttt{AdamW} optimizer~\cite{loshchilov2017decoupled} with $\mathrm{lr}=10^{-5}$; early Stopping (patience $30$ epochs) and \texttt{ReduceLROnPlateau} scheduler (factor $0.1$, patience $10$) monitoring validation \gls{iou} for both phase $1$ and phase 2.
For phase 3, instead, we have reduced the patience of the early stopping to $10$ and for the \texttt{ReduceLROnPlateau} to $4$, again monitoring validation \gls{iou}.

In the preprocessing step (see Figure~\ref{fig:finalmodel}), we applied image augmentations to reduce overfitting and help the model generalize better. 
In particular, we applied translation, rotations ($360$\textdegree), flips, Gaussian noise ($\sigma = 0.01$), and random masking. 
Gaussian noise was introduced to tackle the impact of speckle noise on the segmentation performance. 
After image augmentation, we calculated the prompts related to the current images in the batch as explained in Section~\ref{sec:prompt_robustness}. 
To address class imbalance, we used the Dice loss as it gave us good performance and directly correlated with the \gls{iou} metric. 
Since the mask decoder generates a continuous probability map, a binarization step is required. 
We applied a global threshold of $0.5$, which yielded the best performance in our empirical evaluations, to produce the final segmentation mask.
All models were trained on an \gls{gpu}.

\subsection{Segmentation Tool}
\label{sec:segmentation:tool}
We developed a web-based tool for semi-automated segmentation in collaboration with a geoscientist responsible for the annotations of the \gls{sar} images in the dataset.
The tool is designed to support efficient human-in-the-loop annotation and provides the following core functionalities:
\begin{description}
    \item[Data loading:] loads the files to annotate (\gls{sar} image and the \gls{dem}) at two different time instants $t_0$ and $t_1$.
    \item[Data visualization:] The interface allows visualizing both the RGB composites (obtained from the \gls{sar} images as described in Algorithm~\ref{alg:sar_rgb}) and the \gls{dem}, displayed by simulating a light source to create shadows and highlights (hillshade format), which transforms raw elevation data into a 3D-like representation of the terrain. 
    \item[Semi-automatic segmentation:] the tool allows the annotator to draw a \gls{bb} on the \gls{sar} image around an area with snow avalanche debris. The input data and the prompts are fed to the adapted \gls{sam} that returns a probability map. 
    The annotator can then adjust the mask by setting the threshold for the probability map. The mask corresponding to the selected threshold produces the segmentation mask.
    \item[Mask editing:] enables manual correction/refinement of the mask generated by the deep learning model.
\end{description}

The same web tool can also operate in a fully manual mode, where the annotator draws the avalanches by freehand. 

\begin{figure}
    \includegraphics[width=1\linewidth]{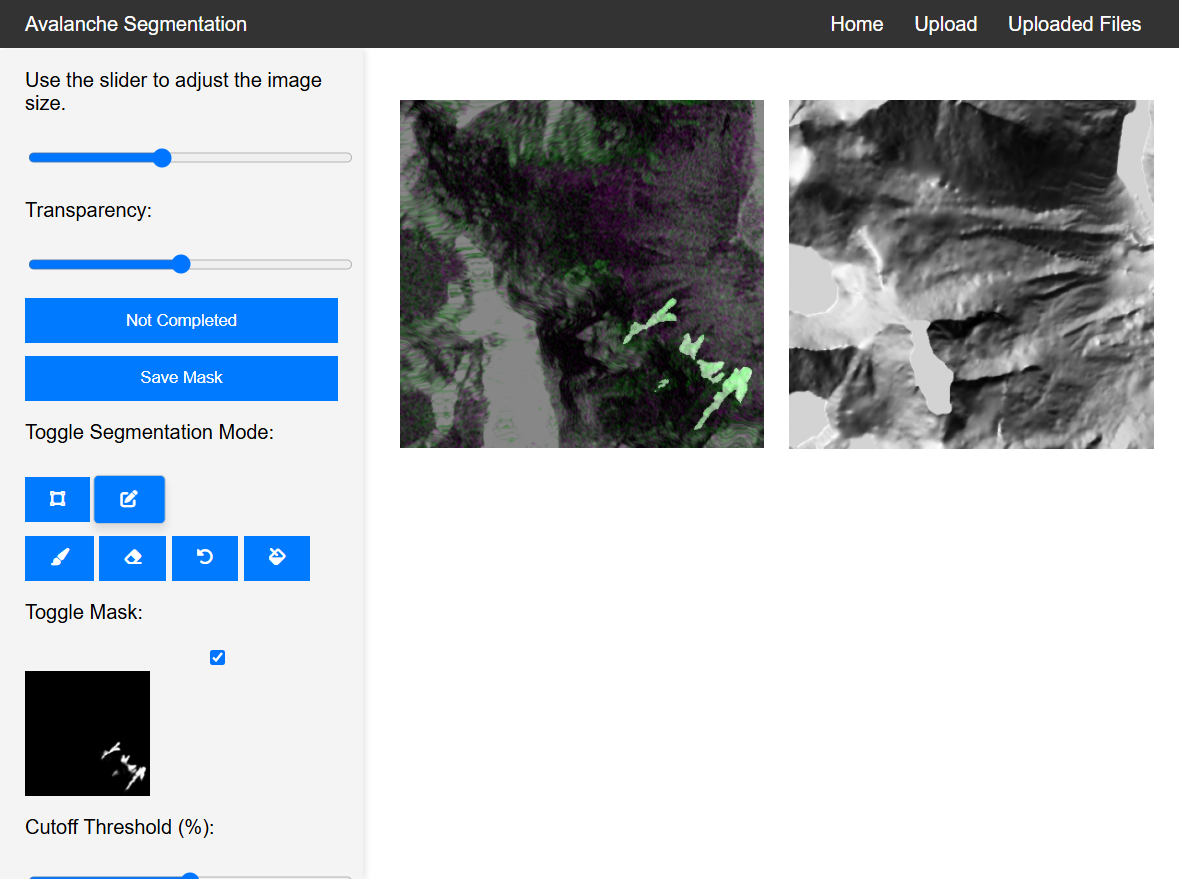}
    \caption{Segmentation tool: here we show the View Image page, which allows for both manual and semi-automatic segmentation of avalanches.}
    \label{fig:viewimage}
\end{figure}
The web page of the software application for image segmentation is shown in Figure~\ref{fig:viewimage}.

\section{Results}
In Section~\ref{sec:ablation_study}, we first conduct an ablation study to assess the benefits provided by each component in our adapted \gls{sam} architecture with respect to baseline methods. 
Then, in Section~\ref{sec:fully_automated} we assess the adapted \gls{sam} model when operating in a fully automatic segmentation and compare it to popular architectures for image segmentation.
Finally, in Section~\ref{sec:seg_tool} we quantitatively assess the practical benefits of our semi-automatic segmentation tool in a real-world annotation pipeline. 
Unless otherwise stated, all models are trained on the avalanche detection dataset using \texttt{VV} and \texttt{VH} \gls{sar} channels along with \gls{dem} and \gls{sa} as inputs. The performance in the experiments is evaluated according to \gls{iou}, precision, and recall, defined in Appendix \ref{sec:figureofmerit}.

\subsection{Ablation Study}
\label{sec:ablation_study}

We compare the effectiveness of the adapted \gls{sam} against:

\begin{itemize}
    \item the zero-shot version of \gls{sam}, taking as input an RGB created through Algorithm \ref{alg:sar_rgb},
    \item the \gls{sam} model from Phase 1 (\gls{sam} with adapters, finetuning of the decoder with \texttt{VV} channels, and the \gls{dem} as input),
    \item the \gls{samm} method \cite{xiao2024segment} with the 6-channel input.
\end{itemize}

We use the same test set with pre-calculated accurate boxes as prompts.
The results are reported in Table \ref{tab:model_comparison}.
Our approach obtains superior performance in almost all metrics, with improvements in \gls{iou} and Recall, which are the most important metrics in avalanche detection.
In particular, our model achieves the highest \gls{iou} metric with respect to any other method we tested.

\begin{table}[!ht]
    \centering
    \caption{Comparison of \gls{sam} adaptation methods for precise prompts}
    \label{tab:model_comparison}
    \begin{tabular}{lccc}
        \toprule
        \textbf{Model} & \textbf{\gls{iou}} & \textbf{Precision} & \textbf{Recall} \\
        \midrule
        \gls{sam} & $34.29$ & $29.51$ & $82.41$ \\
        Phase 1 & $57.88 \pm 0.4$ & $\mathbf{75.79 \pm 1.2}$ & $79.20 \pm 0.8$ \\
        \gls{samm} & $59.17 \pm 0.3$ & $75.58 \pm 0.8$ & $79.97 \pm 1.0$ \\
        Ours & $\mathbf{59.81 \pm 0.3}$ & $75.60 \pm 0.6$ & $\mathbf{80.99 \pm 0.8}$ \\
        \bottomrule
    \end{tabular}
\end{table}

\subsection{Fully-Automatic Segmentation}
\label{sec:fully_automated}

We compare our adapted \gls{sam} in a fully-automatic segmentation, namely providing a full-image box as prompts, against two segmentation baselines fully finetuned on the same multi-channel avalanche dataset: SegFormer-B1 ($13.7M$ parameters)~\cite{paper54} and a U-Net ($14.7M$ parameters)~\cite{paper53}. 

\begin{table}[!ht]
    \centering
    \caption{Comparison of Segmentation Models on Full Image Detection}
    \label{tab:full_detection_comparison}
    \begin{tabular}{lccc}
        \toprule
        \textbf{Model} & \textbf{IoU} & \textbf{Precision} & \textbf{Recall} \\
        \midrule
        U-Net & $42.26 \pm 0.1$ & $\mathbf{68.5 \pm 0.3}$ & $62.73 \pm 0.4$\\
        Segformer & $\mathbf{43.28 \pm 0.7}$ & $68.07 \pm 1.1$ & $65.13 \pm 0.9$\\
        Ours & $42.3 \pm 0.8$ & $62.56 \pm 0.9$ & $\mathbf{66.23
 \pm 1.9}$\\
        \bottomrule
    \end{tabular}
\end{table}

Table \ref{tab:full_detection_comparison} shows that our approach achieves a comparable \gls{iou} and a superior Recall, which is critical to minimize the risk of undetected events.
This indicates the potential to perform highly automated, minimal-prompt segmentation in future applications. 
These results also confirm that the proposed prompt augmentation strategy (described in Section \ref{sec:prompt_robustness}) is effective during training.

We note that to further increase the performance of our \gls{sam} as a fully-automated segmentation model, it would likely require dedicated retraining (e.g., make full-image the priority by increasing their prompt percentage during training).
We also note that in this case, the benefit of using a foundation model like \gls{sam} could be down-weighted by the latency occurring when it is not possible to precalculate the image embedding.
This is closely related to the specific requirements of the application and must be analyzed on a case-by-case basis, depending on the most important performance measure (inference time, precision, \gls{iou}, recall).

\subsection{Semi-automatic Segmentation Tool}
\label{sec:seg_tool}

To evaluate how much the proposed $\text{\gls{sam}}$-based semi-automated segmentation tool (\ref{sec:segmentation:tool}) speeds up the annotation procedure in an operational pipeline, we compared the time required to annotate images in the semi-automatic and in the fully manual mode, within the web tool we developed.
First, we requested an expert geoscientist to generate high-quality annotations for $50$ \gls{sar} images from the test set. 
To perform manual segmentation, the expert took between $1$ and $3$ minutes, while annotations in the sami-automatic modality took about $5$ to $30$ seconds, indicating a substantial speed-up. 

To evaluate if the improvement is statistically significant, we conducted a pairwise comparison through a matched pair analysis on $25$ images, i.e., we tested the significance of the difference in time for segmenting the same image manually or with the automated annotation tool. 
This experiment yielded a $60.28\%$ speedup (using median values), confirmed by a highly significant $p$-value of $10^{-5}$ from the paired one-tailed t-test. 
These outcomes are consistent with similar domain adaptation studies (e.g., $\text{Med-SAM}$~\cite{paper17}), confirming $\text{\gls{sam}}$'s effectiveness in creating segmentation labels in different domains.

\section{Discussion}

This study investigated the adaptation of the \acrfull{sam} framework to snow avalanche segmentation in \acrfull{sar} imagery, with the dual objective of improving segmentation quality and reducing the effort required for manual annotation.
By combining parameter-efficient domain adaptation, prompt-robust training, multi-channel input handling, and compute-aware training, we show that foundation models can be effectively transferred to this highly specialized remote sensing task.

\subsection{Summary of contributions}
The primary contribution of this work is an end-to-end methodology to adapt \gls{sam} to \gls{sar} avalanche data while preserving its prompt-based interaction.
Among the investigated domain adaptation approaches, Adapters proved to be the most effective, enabling efficient fine-tuning by reducing the number of trainable encoder parameters by more than $90\%$.
In addition, the proposed training strategy improves robustness to imprecise prompts, which is essential in realistic human-in-the-loop annotation scenarios.
To overcome the limitation of \gls{sam} to $3$ input channels (Section~\ref{sec:input_adaptation} and Appendix~\ref{app:input_adaptation}), we introduce a multi-encoder architecture based on supervised embedding alignment and fusion, designed to extract complementary information from secondary input channels.
Overall, our final model achieved an \gls{iou} of $0.5981$ using accurate \gls{bb} prompts, representing a $5\%$ improvement over the baseline methods.
When used for fully automatic segmentation, our adapted \gls{sam} model achieved comparable performance to popular image segmentation architectures, namely U-net~\cite{paper53} and Segformer~\cite{paper54} trained end-to-end on the same training set.

The second major contribution is the development of a semi-automatic avalanche annotation tool that, under the hood, runs the proposed \gls{sam}-based segmentation model.
The segmentation tool offers multiple interaction modes, including drawing a \gls{bb} prompt, threshold-based refinement of the returned probability map, and manual mask editing. 
We measured a speed-up of $60.28\%$ compared to the manual annotation process, directly addressing the main bottleneck for scaling up avalanche inventories.
This tool has significant implications for operational avalanche monitoring systems, where timely and accurate detection is critical for public safety. 
Indeed, facilitating the annotation procedure could enable avalanche forecasting centers to process significantly larger volumes of \gls{sar} imagery, potentially improving the temporal and spatial coverage of avalanche monitoring programs. 
This scalability is particularly relevant given the increasing availability of \gls{sar} data from missions such as Sentinel-1, which provides regular coverage of mountainous regions regardless of weather conditions.

More broadly, increasing the amount of high-quality labels enables training more accurate and reliable models for snow avalanche detection, while larger and more diverse datasets remain the main bottleneck for automated snow avalanche mapping.
In the long term, the proposed tool can support a positive feedback loop in which improved models reduce annotation effort and facilitate further dataset expansion.

\subsection{Challenges and limitations}
One of the main technical challenges faced during this work was the long training time, which we addressed with the proposed efficient training algorithm.
Looking ahead, the most important performance bottleneck concerns the detection of small avalanches: besides the inclusion of imprecise prompts, the other solutions we tested did not consistently improve segmentation performance (Appendix~\ref{small_optimization}).

Another limitation is that the dataset includes only acquisitions from the Norway region. 
The generalization of our model to other geographic areas with different snow conditions and terrain characteristics is not guaranteed and requires additional validation. 
Nevertheless, we argue that the proposed methodology is general, and the same architecture could be kept as-is and retrained on data from different regions.

We also acknowledge that the simpler adapter model with three input channels created through Algorithm~\ref{alg:sar_rgb} represents a strong baseline in terms of \gls{iou}.
Nevertheless, the proposed multi-encoder procedure provides a principled way to incorporate additional channels when they are available and informative for the downstream task.

\subsection{Conclusions and future work}
Snow avalanche mapping in \gls{sar} imagery is inherently challenging due to speckle noise, acquisition timing, and inter-annotator variability.
Our results show that a promptable foundation model, once properly adapted, can act as an effective assistant for this task: it produces accurate masks from simple \glspl{bb} and, when integrated in an operational pipeline, substantially reduces annotation time.
This creates a practical opportunity to scale up high-quality avalanche inventories, which can in turn improve both prompt-based and fully automatic detection systems.

Beyond avalanche mapping, the proposed methodology for multi-modal input handling and prompt-based training is applicable to other \gls{sar}-based detection tasks, including flood detection~\cite{amitrano2024flood} and oil spill detection~\cite{bianchi2020large}.

Future research should prioritize the following key directions:
\begin{itemize}
    \item expand training data through larger annotation campaigns supported by the tool, with quality-control protocols that improve contour consistency across annotators.
    \item investigate multi-scale decoding and training strategies that increase sensitivity to small targets while controlling false positives.
    \item validate and retrain on acquisitions from other regions and seasons, and explore additional inputs when available (e.g., meteorological products or higher-resolution topographic descriptors).
    \item conduct field evaluations with forecasting centers to assess usability, latency, and reliability in real annotation workflows.
\end{itemize}
}

\providecommand{\appendixtitles}[1]{}
\providecommand{\appendixstart}{}
\providecommand{\appendixname}{Appendix}

\newif\ifisarxiv

\newcommand{\PaperAppendix}{%
\newpage

\ifisarxiv
\section*{Appendix}
\fi

\appendixtitles{no} 
\appendixstart
\appendix
\section[\appendixname~\thesection]{Evaluation Metrics}
\label{sec:figureofmerit}

\gls{iou} represents the main metric and determines how closely the predicted avalanche area matches the area in the ground truth. The \gls{iou} is defined as:
\begin{equation}
    \text{IoU} = \frac{|A \cap B|}{|A \cup B|}
\end{equation}
Where $A$ is the predicted mask and $B$ is the ground truth annotation.

The precision is defined as:
\begin{equation}
    \text{Precision} = \frac{\text{TP}}{\text{TP} + \text{FP}}.
\end{equation}
Where TP are the True Positives, FP the False Positives.
Precision determines the percentage of predicted avalanche pixels that were actually correct, and it is crucial in cases where false alarms carry high operational costs or severe consequences.

The recall is defined as:
\begin{equation}
    \text{Recall} = \frac{\text{TP}}{\text{TP} + \text{FN}}.
\end{equation}
Where FN are the False Negatives Recall determines the percentage of actual avalanche pixels that the model successfully identifies, and it is crucial in safety-critical applications where missing a detection can have severe consequences.
Recall is an important metric for fully autonomous avalanche detection systems, where missing avalanches can have significant impacts and endanger lives.
It is often desirable to avoid missing potentially hazardous regions at the cost of making the model more trigger-happy.
\section[\appendixname~\thesection]{Input Adaptation}
\label{app:input_adaptation}

\subsection[\appendixname~\thesubsection]{Standard RGB creation for manual segmentation}
\label{manual_segmentation}

To perform manual segmentations, expert annotators rely on two different types of RGB composites.
The first is simply obtained by combining the following channels: $[\texttt{VV}_0, \texttt{VV}_1, \texttt{VV}_0]$.
The other RGB composite is obtained through Algorithm \ref{alg:sar_rgb}.

\begin{algorithm}
\caption{\gls{sar} Polarimetric Data to RGB Image Conversion}
\label{alg:sar_rgb}
\begin{algorithmic}[1]
\Function{CreationOfRGBImage}{$\texttt{VH}_0, \texttt{VH}_1, \texttt{VV}_0, \texttt{VV}_1$}
    \State $\texttt{VH}_i \leftarrow \texttt{rescale}(\texttt{VH}_i, -27, -7)$ for $i = 0,1$ \label{line:sar1}
    \State $\texttt{VV}_i \leftarrow \texttt{rescale}(\texttt{VV}_i, -23, -3)$ for $i = 0,1$ \label{line:sar2}
    \State $a \leftarrow \texttt{rescale}(\texttt{VH}_1 - \texttt{VH}_0, 0, 0.25)$ \label{line:sar3}
    \State $b \leftarrow \texttt{rescale}(\texttt{VV}_1 - \texttt{VV}_0, 0, 0.25)$ \label{line:sar4}
    \State $w \leftarrow \texttt{rescale}(a - b, 0, 1)$ \label{line:sar5}
    \State $R \leftarrow w \cdot \texttt{VH}_0 + (1 - w) \cdot \texttt{VV}_0$ \label{line:sar6}
    \State $G \leftarrow w \cdot \texttt{VH}_1 + (1 - w) \cdot \texttt{VV}_1$ \label{line:sar7}
    \State $B \leftarrow w \cdot \texttt{VH}_0 + (1 - w) \cdot \texttt{VV}_0$ \label{line:sar8}
    \State $\text{RGB} \leftarrow [R, G, B]$ \label{line:sar9}
    \State \Return{RGB}
\EndFunction
\end{algorithmic}
\end{algorithm}

The function \texttt{rescale} converts the \gls{sar} data, which is usually provided in logarithmic scale (dB), to the interval $[0,1]$ to enhance the effectiveness of the detection algorithms.
Algorithm \ref{alg:rescale_rgb} shows the details of the \texttt{rescale} function, where ``arr'' represents the input data, while ``lo'' and ``hi'' are user-defined thresholds.

\begin{algorithm}
\caption{Rescale Algorithm}
\label{alg:rescale_rgb}
\begin{algorithmic}[1]
\Function{rescale}{$\text{arr}, \text{lo}, \text{hi}$}
    \State $\text{arr} \leftarrow \frac{\text{arr} - \text{lo}}{\text{hi} - \text{lo}}$
    \State $\text{arr} \leftarrow 0$ if \text{arr} < 0
    \State $\text{arr} \leftarrow 1$ if \text{arr} > 1
    \State $\text{arr} \leftarrow 0$ if \text{isnan}(\text{arr})
    \State \Return $\text{arr}$
\EndFunction
\end{algorithmic}
\end{algorithm}

After the initial rescaling of the raw \gls{sar} data (Lines \ref{line:sar2}-\ref{line:sar3} of Algorithm \ref{alg:sar_rgb}), we calculate the difference between time steps $t_0$ and $t_1$, and we rescale it again between $0$ and $1$ (Lines \ref{line:sar4}-\ref{line:sar5}).
The pixel values in the difference image will be higher where a new avalanche occurred, due to the higher diversity in the back-scattering. 
We then subtract the two difference images $a$ and $b$, obtaining a new output $w$ (Line \ref{line:sar6}), which will be $0$ where the backscattering difference is higher in the \texttt{VV} image and a value between $0$ and $1$, otherwise.
We then use $w$ for a convex combination of \texttt{VV} and \texttt{VH} (Lines \ref{line:sar7}-\ref{line:sar9}).

Overall, the \texttt{VV} polarization is more informative for avalanche detection. This behavior is reflected in Algorithm~\ref{alg:sar_rgb}: the weight $w$ is typically small, so the resulting RGB composite is dominated by \texttt{VV} and the influence of \texttt{VH} is reduced. The \texttt{VH} channel becomes informative only where its backscatter change exceeds that of \texttt{VV}; even then, its contribution to the final image remains minor.

We also underline that, in addition to the RGB composites, the annotator relies on the topographic information contained in the \gls{dem} or, more precisely, in products obtained from the \gls{dem}: the hill-shade representation (which is a better way to visualize the topology) and the \gls{sa}, which gives information about the slope and steepness.

\subsection[\appendixname~\thesubsection]{Channels combination}
\label{channels_combination}

\begin{figure}[H]
    \centering 
    
    \subfloat[Standard]{\includegraphics[width=0.31\textwidth]{Images/imageAvalanche2.png}}
    \hspace{0.1cm}
    \subfloat[Standard + \gls{dem}]{\includegraphics[width=0.31\textwidth]{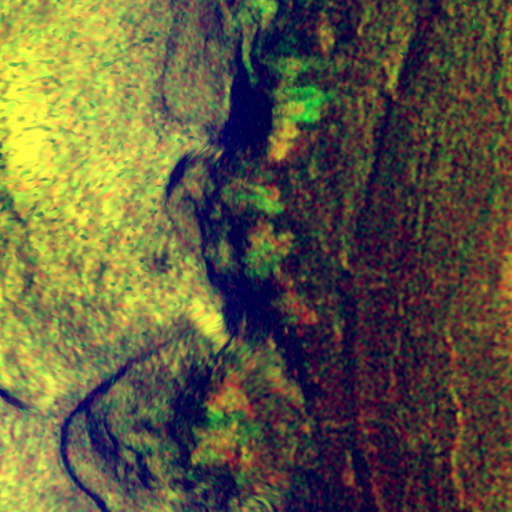}}
    \hspace{0.1cm}
    \subfloat[Difference  + \gls{dem}]{\includegraphics[width=0.31\textwidth]{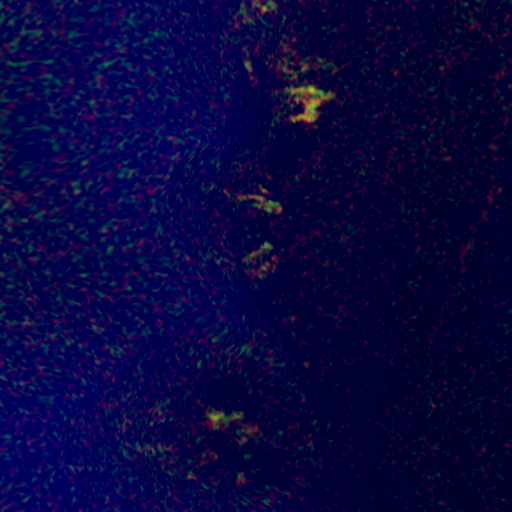}}
    
    \subfloat[Vertical  + \gls{dem}]{\includegraphics[width=0.31\textwidth]{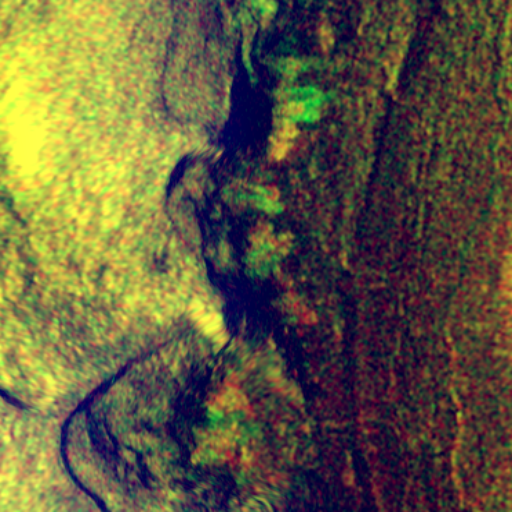}}
    \hspace{0.1cm}
    \subfloat[Horizontal  + \gls{dem}]{\includegraphics[width=0.31\textwidth]{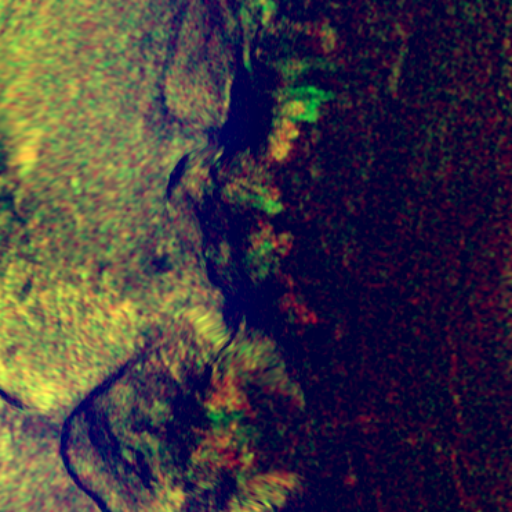}}
    \hspace{0.1cm}
    \subfloat[Ground Truth]{\includegraphics[width=0.31\textwidth]{Images/imageMask2.png}}
    
    \caption{Visual comparison of different \gls{sar} input modalities and feature combinations: 
    (\textbf{a}) The RGB composite for manual interpretation from Algorithm \ref{alg:sar_rgb}; 
    (\textbf{b}) The RGB composite of \textbf{a} combined with the \gls{dem} substituted in the third channel; 
    (\textbf{c}) creates an RGB with [$\texttt{VV}_1$ - $\texttt{VV}_0$, $\texttt{VH}_1$ - $\texttt{VH}_0$, \gls{dem}]; 
    (\textbf{d}) creates an RGB with [$\texttt{VV}_0$, $\texttt{VV}_1$, \gls{dem}]; 
    (\textbf{e}) creates an RGB with [$\texttt{VH}_0$, $\texttt{VH}_1$, \gls{dem}]; 
    (\textbf{f}) reference ground truth mask for avalanche segmentation.}
    \label{fig:combinations}
\end{figure}

A straightforward unimodal way to adapt \gls{sam} to avalanche segmentation is to rescale the selected modalities and stack them into a three-channel (RGB) input.
Figure~\ref{fig:combinations} shows examples of the considered configurations.
Using the standard channels ($\texttt{VV}_0$, $\texttt{VV}_1$, $\texttt{VH}_0$, $\texttt{VH}_1$, \gls{dem}, \gls{sa}), the simplest composites are obtained by rescaling and directly stacking:

\begin{itemize}
    \item Vertical  + \gls{dem}: creates an RGB with [$\texttt{VV}_0$, $\texttt{VV}_1$, \gls{dem}].
    \item Horizontal  + \gls{dem}: creates an RGB with [$\texttt{VH}_0$, $\texttt{VH}_1$, \gls{dem}].
\end{itemize}

We also considered more complex variants:

\begin{itemize}
    \item Difference  + \gls{dem}: creates an RGB with [$\texttt{VV}_1$ - $\texttt{VV}_0$, $\texttt{VH}_1$ - $\texttt{VH}_0$, \gls{dem}].
    \item Standard  + \gls{dem}: creates an RGB with [R, G, \gls{dem}], where $R$ and $G$ are computed as in Algorithm~\ref{alg:sar_rgb}.
\end{itemize}

Table~\ref{tab:inputcomparison} summarizes the performance obtained when using as input the different combinations of channels.

\begin{table}[h!]
    \caption{Performance Metrics (\gls{iou}, Precision, Recall) for different input configurations}
    \label{tab:inputcomparison}
    
    \centering 
    \setlength{\tabcolsep}{10pt} 

    \begin{tabular}{l c c c}
        \toprule
        \textbf{Input Configuration} & \textbf{IoU} & \textbf{Precision} & \textbf{Recall} \\
        \midrule
        Vertical  + \gls{dem} & 58.21 & 77.18 & \textbf{78.38} \\ 
        Horizontal  + \gls{dem} & 54.58 & 71.88 & 78.30 \\
        Difference  + \gls{dem} & 57.39 & 75.68 & 77.77 \\
        Standard  + \gls{dem} & \textbf{59.16} & \textbf{79.08} & 77.78 \\ 
        \bottomrule
    \end{tabular}
\end{table}

These unimodal three-channel composites already provide competitive performance: the best configuration (Standard  + \gls{dem}) reaches an \gls{iou} of $59.16$, which is close to the final results reported in the main body (Table~\ref{tab:model_comparison}). 
This supports the discussion in the main text that careful input selection yields a strong starting point, and that more complex multi-modal strategies provide a modest but meaningful additional gain. Moreover, our final model is worse in terms of precision and better in terms of recall. Overall, the final model outperforms this best unimodal approach in the key metrics in avalanche segmentation. Considering the presence of false negatives in the labels used for training, it is better to have a less conservative model, which could reduce this problem.

\subsection[\appendixname~\thesubsection]{Patch Embedding}
\label{patch_embedding}

\gls{sam}'s patch embedding is the first layer of the image encoder and the only component whose parameters depend on the number of input channels. 
Replacing the patch embedding layers is, therefore, the most direct way to ingest all available modalities. 
However, this solution also discards the pretrained projection that maps the input image to the token embeddings expected by the subsequent \gls{vit} blocks, which can negatively affect optimization and downstream components.

We experimented to directly embed the 6-channel input ($\texttt{VV}_0$, $\texttt{VV}_1$, $\texttt{VH}_0$, $\texttt{VH}_1$, \gls{dem}, \gls{sa}) and trained \gls{sam} with the new, initialized, patch embedding. 
As expected, this model did not achieve the same performance as the strongest unimodal baselines, reaching a maximum \gls{iou} of $57.17$.

To mitigate the disadvantage of losing the pretrained parameters, we investigated self-supervised initialization strategies for the new patch embedding (while also training the adapters):

\begin{itemize}
    \item \textbf{Masked reconstruction.} We trained the modified encoder on unlabeled data to reconstruct the input from a masked version (masking ratio $30\%$). To limit training time and encourage the encoder to carry most of the representation burden, we used a lightweight decoder composed of three transposed-convolution layers.
    \item \textbf{Embedding distillation.} We extracted image embeddings with the original \gls{sam} from a 3-channel input and trained the modified model to reproduce the same embeddings starting from the 6-channel input.
\end{itemize}

After self-supervised pretraining, supervised fine-tuning improved the results to $57.73$ and $58.63$ \gls{iou}, respectively, with the embedding-distillation objective providing the largest gain. 
We hypothesize that the unsupervised pretraining stabilizes optimization by improving gradient flow and by reducing the mismatch between the encoder output and the embedding space expected by the decoder. 
Nevertheless, the best patch-embedding variant remained below the strongest three-channel configuration (Table~\ref{tab:inputcomparison}) and was therefore not retained in the final model.

\subsection[\appendixname~\thesubsection]{Prefix Net}
\label{prefix_net}

\begin{figure}
    \includegraphics[width=1\linewidth]{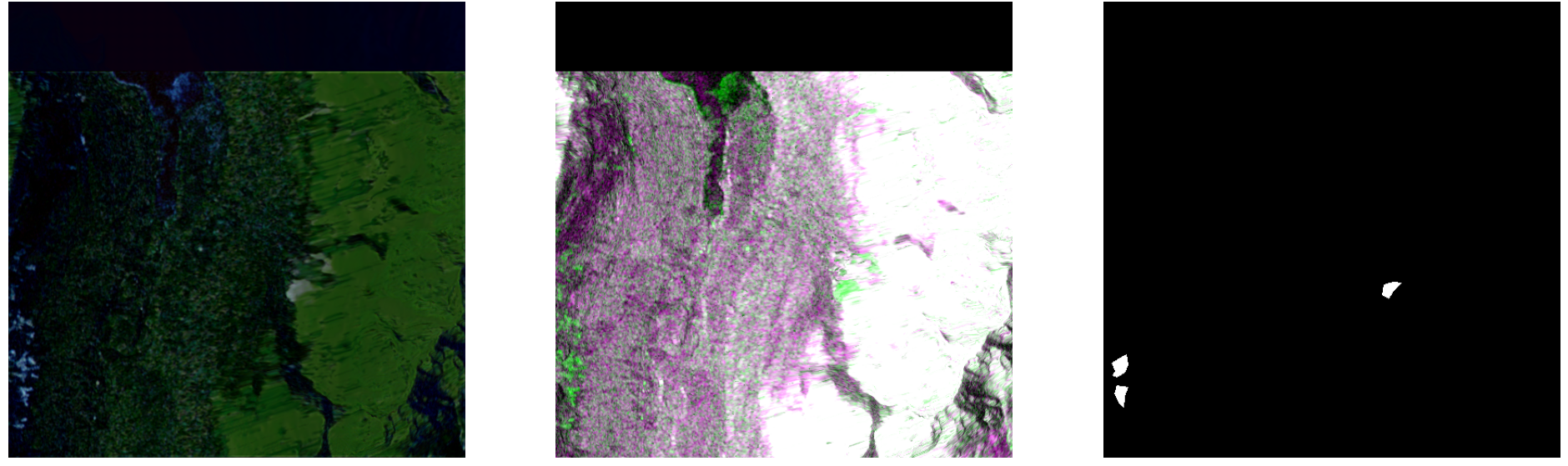}
    \caption{Left: Output of the Prefix Net. Center: standard image used to perform manual segmentation by experts. Right: the ground-truth mask. As we can see, the avalanches are more clearly visible on the output of the Prefix Net.}
    \label{fig:prefixoutput1}
\end{figure}

Instead of modifying the pretrained image encoder, the Prefix Net approach learns a mapping from an $n$-channel input to a 3-channel pseudo-RGB image, which is then fed to the original \gls{sam} patch embedding and encoder. 
This can be seen as a data-driven alternative to the hand-engineered channel combinations described in Appendix~\ref{channels_combination}.

We considered two variants:
\begin{itemize}
    \item \textbf{Light:} a small convolutional network designed for efficient end-to-end training.
    
    The input is first processed by two $3 \times 3$ convolutional layers with Batch Normalization. This block maintains the input resolution while projecting the features into a $64$-channel latent space. A Transpose Convolutional layer (ConvTranspose2d) with a $2 \times 2$ kernel and a stride of 2 increases the resolution to $1024 \times 1024$. The final stage consists of a ReLU activation followed by a $3 \times 3$ convolutional layer that maps the $64$ intermediate features to $3$.
    \item \textbf{Heavy:} a multi-branch convolutional architecture with parallel kernels to better capture multi-scale patterns and improve small-avalanche segmentation inspired by Dong \emph{et al.} \cite{paper5}.
\end{itemize}

End-to-end training of the heavy variant exceeded our GPU memory budget. 
We therefore trained the prefix network first with the original \gls{sam} and then fine-tuned \gls{sam} (with adapters) using the pretrained prefix output. 
This two-stage procedure reached $54.45$ \gls{iou}, substantially below the light variant and below the best unimodal baselines. 
We expect that joint or alternating optimization could be more effective given sufficient hardware resources.

The light Prefix Net can be trained jointly with the adapters and achieved an \gls{iou} of $59.65$, slightly outperforming the best hand-crafted three-channel composite. 
It also reduces the negative correlation between avalanche size and \gls{iou}, improving performance on small avalanches. 
In addition, the learned mapping tends to denoise the inputs and produces visually interpretable 3-channel images (Fig.~\ref{fig:prefixoutput1}), which can support manual inspection and annotation, as discussed in Section~\ref{sec:input_adaptation}.

\section[\appendixname~\thesection]{Self-Supervised Pretraining}
\label{self-supervised_pretraining}

As discussed in Section~\ref{sec:domain_adapt}, we adapt \gls{sam} to the \gls{sar} avalanche domain by fine-tuning lightweight adapter modules while keeping the pretrained image-encoder backbone frozen. Since the adapters are randomly initialized, we explored self-supervised pretraining as a way to provide a more informative initialization, improve training stability, and potentially learn noise-robust representations for \gls{sar} imagery.

We tested multiple self-supervised initialization methods, summarized here for completeness:

\begin{itemize}
    \item Masked-autoencoders: this is the standard method of pretraining for \glspl{vit}, which reconstructs the input from a masked version \cite{paper33}.
    \item Teacher-student: learn view-invariant representations by matching student and teacher embeddings across multiple augmentations. 
    \item Self-supervised denoising: reconstruct a filtered target to encourage noise-robust features.
\end{itemize}

Since annotations are not required, it was possible to collect additional data from Norway and obtain an independent dataset for the self-supervised training, which is substantially larger than the supervised avalanche dataset (over $10,000$ images for training and over $1,000$ for validation). 
The larger dataset implied a significantly higher computational cost: in our setting, each pretraining run required several days to converge.

\subsection[\appendixname~\thesubsection]{Masked-autoencoders}
We follow the \gls{mae} paradigm \cite{paper33}. 
During pretraining, we keep the pretrained \gls{sam} image encoder frozen and train only the adapters together with a lightweight reconstruction head.
The head maps the image embedding back to the input space and is implemented as a small convolutional decoder with three layers (kernel size $3$). 
A lightweight decoder is commonly preferred in \gls{mae}-style training because it encourages the encoder to learn informative representations rather than delegating reconstruction capacity to the decoder.

We use as input the standard 3-channel composite produced by Algorithm~\ref{alg:sar_rgb}. 
The masking ratio is set to $30\%$ (lower than the typical $70\%$ used in natural-image \gls{mae}), motivated by the lower and noisier information content of \gls{sar} composites. 
On the \gls{gpu}, training converged in about four days with early stopping (patience $20$ epochs). 
After pretraining, we discard the reconstruction head and fine-tune the model for avalanche segmentation using the baseline supervised protocol (adapters and decoder unfrozen). 
This initialization reached a best \gls{iou} of $58.63$, which is below the performance obtained with randomly initialized adapters.

\subsection[\appendixname~\thesubsection]{Teacher-student}

We also explored a DINO-style teacher-student training procedure \cite{paper31}. 
We use \gls{sam} with adapters as the backbone for both teacher and student, initialized with identical weights. 
During self-supervision, only the adapters are trained; the teacher is updated through an exponential moving average of the student parameters.

Due to GPU memory constraints ($48$GB VRAM), we use only two global views and do not include local crops. 
The global views are generated through standard image augmentations (rotation, translation, flips, Gaussian blur, grayscale, random crops, and color jitter). 
Training converged in about four days on the \gls{gpu}, with early stopping (patience $20$ epochs). 
Fine-tuning the student initialization for avalanche segmentation yielded a best \gls{iou} of $58.73$, which again did not improve over the baseline with random adapter initialization.

\subsection[\appendixname~\thesubsection]{Self-supervised denoising}
\begin{figure}
    \centering 
    
    \subfloat[Original Image Image]{\includegraphics[width=0.31\textwidth]{Images/imageSARDEM.png}}
    \hspace{0.1cm}
    \subfloat[Lee Filtered]{\includegraphics[width=0.31\textwidth]{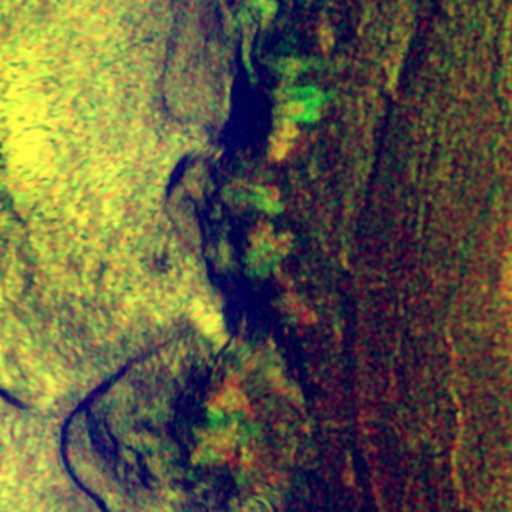}}
    \hspace{0.1cm}
    \subfloat[Edge Detection Kernels]{\includegraphics[width=0.31\textwidth]{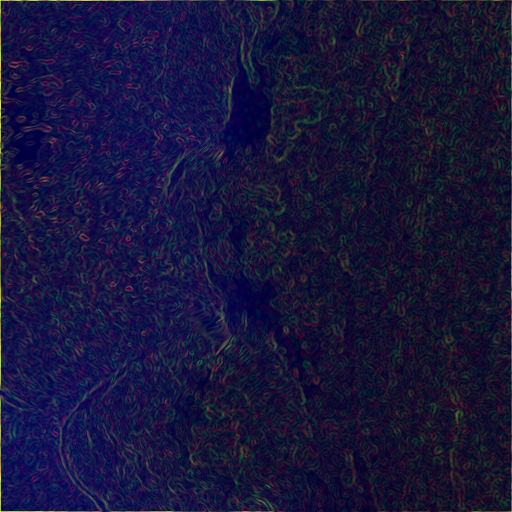}}
    
    \subfloat[Gradient Transform]{\includegraphics[width=0.31\textwidth]{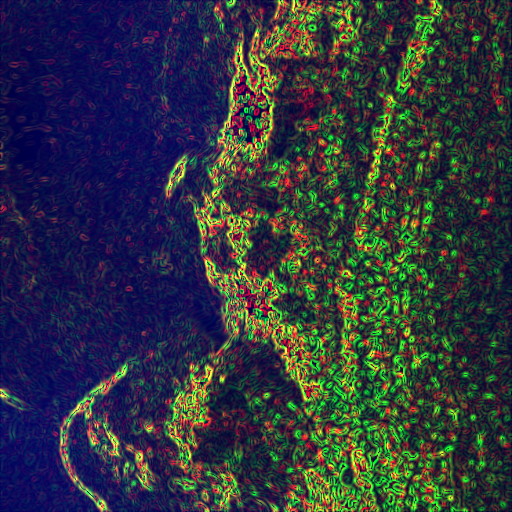}}
    \hspace{0.1cm}
    \subfloat[Ground Truth]{\includegraphics[width=0.31\textwidth]{Images/imageMask2.png}}
    
    \caption{Examples of denoising targets considered for self-supervision: (\textbf{a}) input composite; (\textbf{b}) Lee-filtered target; (\textbf{c}) edge-based target; (\textbf{d}) gradient-based target; (\textbf{e}) reference ground truth mask.}
    \label{fig:feature_grid}
\end{figure}

Motivated by prior work on self-supervised denoising for \gls{sar} data \cite{paper22, paper24}, we consider reconstruction tasks in which the target is not the raw input but a denoised or feature-enhanced version.
We use \gls{sam} with adapters (only adapters trainable) and attach a lightweight fully convolutional decoder (three layers) to reconstruct the target.
As in \gls{mae}, $30\%$ of the input is masked. 
In addition, when the $3$-channel input includes a topographic channel, we reconstruct \gls{sa} instead of \gls{dem}, forcing the model to preserve terrain-relevant cues.

We evaluated multiple target transformations, including the gradient-based representation proposed in \cite{paper22}, a Lee filter for speckle reduction, and an edge-based target inspired by IRSAM \cite{paper15}.
Figure~\ref{fig:feature_grid} shows representative outputs. 
In our data, the gradient- and Lee-based targets tended to suppress or blur avalanche boundaries, whereas edge-based targets provided a simpler objective that better preserves boundary information. 
We therefore adopted edge reconstruction (for the \gls{sar} channels) together with \gls{sa} reconstruction as our denoising pretext task.

After supervised fine-tuning, this initialization achieved a best \gls{iou} of $58.89$, which was the strongest among the self-supervised objectives but still below the baseline with random adapter initialization. 
Given the additional training time, we did not include this strategy in the final model.

\subsection[\appendixname~\thesubsection]{Conclusions on Model Initialization}

Overall, self-supervised pretraining did not improve the downstream segmentation performance in our setting.
We attribute this outcome to a mismatch between the image-embedding distribution induced by the self-supervised objectives and the embedding space expected by the pretrained \gls{sam} mask decoder. 
In practice, this manifests as an unstable initial fine-tuning phase that is not fully compensated during supervised training, even when the learned features appear qualitatively meaningful.

We found that freezing the image encoder for the first 10 epochs of supervised fine-tuning (allowing the decoder to adapt to the pretrained embeddings) improves stability; the results reported above include this additional step. 
Nevertheless, none of the tested objectives consistently outperformed the baseline with randomly initialized adapters, and we therefore adopted standard initialization in the final approach.

\section[\appendixname~\thesection]{Meteorological Data}
\label{meteo_data}
Meteorological conditions play an important role in avalanche release and are widely used in operational forecasting \cite{paper3, paper2}.
In our dataset, each image is associated with $5$ time series, each one measuring a different meteorological variable (temperature, wind speed, air pressure, precipitation amount, relative humidity), described in Table \ref{tab:met_stats}.

\begin{table}[ht]
    \centering
    \caption{Descriptive Statistics for Meteorological Variables}
    \begin{tabular}{lccc}
        \toprule
        Variable & Mean & Std & Unit \\
        \hline
        Air Temperature (2m) & 270.59 & 3.74 & K \\
        Wind Speed (10m) & 7.1 & 4.42 & m/s \\
        Air Pressure at Sea Level & 100348.15 & 1650.62 & hPa \\
        Precipitation Amount & 0.274 & 0.68 & mm \\
        Relative Humidity (2m) & 0.86 & 0.135 & \% \\
        \bottomrule
    \end{tabular}
    \label{tab:met_stats}
\end{table}

\gls{met} data are characterized in out dataset by an extremely low spatial resolution.
In particular, there is only a time series associated with each \gls{sar} image, which spans the entire duration $[t_0, t_1]$ between the two satellite passes with a time resolution of 1 hour.
Since \gls{sam} operates on spatial prompts (Figure~\ref{fig:sam}), using these time series requires converting them into a dense, image-aligned representation. 
To provide the spatial structure, we condition the meteorological embedding on topography by pairing it with the \gls{sa}. 
Inspired by MetNet2 \cite{paper48}, we encode the normalized slope map with a small convolutional block (\texttt{Conv}-\texttt{LayerNorm}-\texttt{ReLU}), and we encode the normalized meteorological sequences with an LSTM \cite{hochreiter1997long} to capture temporal dependencies. 
The LSTM output is projected to a feature vector, broadcast over the spatial grid, concatenated with the slope features, and processed with additional convolutional blocks to produce the final dense prompt encoding.

We trained this meteorology-conditioned prompt jointly with the adapter baseline, enabling it for 50\% of the samples for which \gls{met} data are available. 
At test time, however, using the prompt degraded performance (\gls{iou} $58.12$) compared to disabling it (\gls{iou} $59.06$). 
We attribute this to the very coarse spatial granularity of the meteorological measurements and to the fact that, in our setting, they do not provide discriminative information at the pixel level beyond what is already captured by \gls{sar} and topography. 
For these reasons, meteorological data are not included in the final model.

\section[\appendixname~\thesection]{Small Avalanche Optimization}
\label{small_optimization}
\begin{figure}
    \centering
    \includegraphics[width=0.9\linewidth]{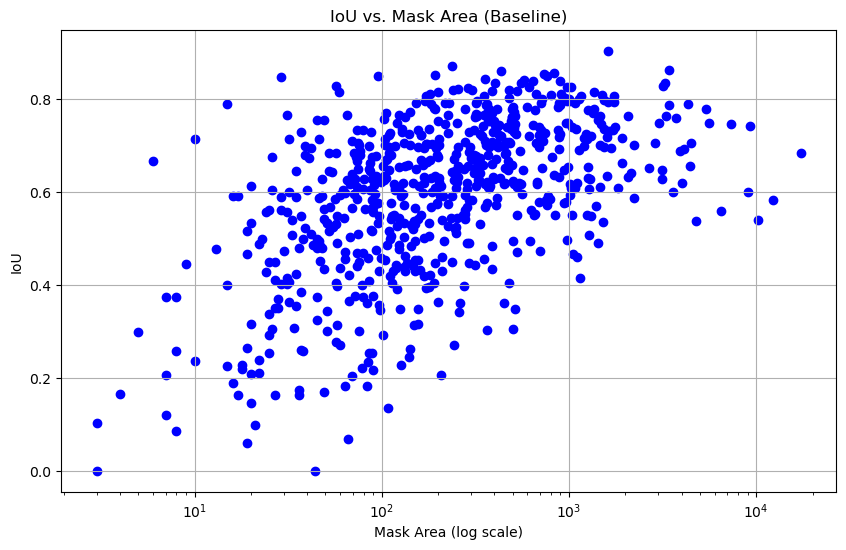}
    \caption{Scatter-plot of the \gls{iou} versus the mask area in log scale. The clear correlation suggests that smaller snow avalanches are harder to detect.}
    \label{fig:iou-size-corr}
\end{figure}
As discussed in Section \ref{sec:prompt_robustness}, segmentation performance decreases for small avalanches, leading to a positive correlation between avalanche area and \gls{iou} (see Figure \ref{fig:iou-size-corr}).
Besides the \gls{bb} augmentation strategy adopted in the main body in Section \ref{sec:prompt_robustness}, here evaluated in Table \ref{tab:correlation} on the model with Adapters and finetuning of the decoder with the image created by Algorithm \ref{alg:sar_rgb} as input. 
\begin{table}[H]
  \centering
  \caption{Correlation Coefficient and p-value for Variable Pairs Based on Prompt Training Strategy}
  \label{tab:correlation}
  \begin{tabular}{lccc}
    \hline
    Variable Pair & Prompt & Correlation Coefficient ($r$) & p-value \\
    \hline
    IoU vs Mask Area & Accurate Only & $0.2056$ & $4.86\times 10^{-8}$ \\
    {} & Ours & $0.1668$ & $1.06\times 10^{-5}$ \\
    \hline
  \end{tabular}
\end{table}
We evaluated additional approaches commonly used to improve the detection of small targets:

\begin{itemize}
    \item \textbf{Loss reweighting:} modify the objective to emphasize hard examples and mitigate class imbalance (e.g., Dice and Focal losses) \cite{paper1, lin2017focal}.
    \item \textbf{Multi-scale feature extraction:} combine features computed at different receptive fields to capture better small structures and sharp boundaries \cite{paper5}.
    \item \textbf{High-resolution feature injection:} add skip connections so that the decoder can leverage earlier, higher-resolution features, as in U-Net-like architectures \cite{paper,paper6, paper53}.
\end{itemize}

Regarding loss reweighting, we use Dice as the default objective because it addresses class imbalance and directly correlates with \gls{iou}. 
We also tested Dice+Focal, which is frequently used for small or hard-to-classify targets \cite{paper49, paper50}, and class-weighted losses (doubling the avalanche weight following \cite{paper}). 
In our experiments, these variants performed worse than Dice alone and were not pursued further.

Multi-scale and skip-connection mechanisms are not directly available in \gls{sam}, whose image encoder is a \gls{vit} and whose convolutional components appear only in the patch embedding and in the final stages of the encoder/decoder. 
An Auto-SAM-style solution is to add an auxiliary convolutional branch in parallel to \gls{sam} \cite{paper11}. 
We also investigated skip-connection variants that inject higher-resolution features into the decoder.

In particular, we evaluated HQ-SAM \cite{paper14} and IRSAM \cite{paper15}, which expose the decoder to intermediate encoder features. 
These approaches achieved \gls{iou} values of $59.21$ and $59.02$, respectively. 
The gains were marginal and did not justify the added architectural complexity and memory overhead.

A limitation of HQ-SAM is that the connected features originate from global-attention blocks (the first connection is after $3$ of $12$ transformer blocks), where representations are already highly processed. 
The added convolutional adapters must simultaneously preserve information for the remaining transformer blocks and provide high-frequency signals for the skip connection. 
To provide less processed, higher-resolution signals, we implemented an additional convolutional branch that feeds features directly from the input to the mask decoder using the HQ-SAM fusion mechanism. 
The tested branch consists of:

\begin{itemize}
    \item \textbf{Initial block:} strided $3 \times 3$ convolutions reducing $1024 \times 1024 \times 3$ to $512 \times 512 \times 64$, followed by \texttt{BatchNorm} and \texttt{ReLU}.
    \item \textbf{Middle block:} three residual stages with $3 \times 3$ convolutions and max pooling, reducing to $64 \times 64 \times 256$ (with a $1 \times 1$ projection in the identity path when channel dimensions change).
    \item \textbf{Final block:} a linear convolution producing the features injected into the decoder.
\end{itemize}

This variant reached a best \gls{iou} of $59.39$ in early experiments. 
However, the improvement was not reproducible once we introduced the less precise prompt strategy and the multi-encoder components of the final method. After careful evaluation, we found that these architectural and loss-function variants were less effective for small-avalanche performance than the \glspl{bb} augmentation procedure described in Section~\ref{sec:prompt_robustness}.
}

\isarxivtrue

\algnewcommand\Input{\item[\textbf{Input:}]}
\algnewcommand\Output{\item[\textbf{Output:}]}

\title{Promptable Foundation Models for SAR Remote Sensing: Adapting the Segment Anything Model for Snow Avalanche Segmentation}

\author{%
Riccardo Gelato$^{1}$ \And
Carlo Sgaravatti$^{1}$ \And
Jakob Grahn$^{2}$ \And
Giacomo Boracchi$^{1}$ \And
Filippo Maria Bianchi$^{2,3}$\thanks{Correspondence: filippo.m.bianchi@uit.no}\\[0.75em]
$^{1}$ DEIB - Dipartimento di Elettronica Informazione e Bioingegneria, Politecnico di Milano\\
$^{2}$ NORCE Norwegian Research Centre AS\\
$^{3}$ UiT The Arctic University of Norway%
}

\begin{document}
\maketitle

\begin{abstract}
\PaperAbstract
\end{abstract}

\PaperBody

\section*{Acknowledgments}
F. M. B. is supported by the Norwegian Research Council project no. 345017~(\textit{RELAY:~Relational Deep Learning for Energy Analytics}).
The authors wish to thank Nvidia Corporation for donating the GPUs used in this project.

\PaperAppendix

\bibliographystyle{unsrtnat}
\bibliography{bibliography}

@article{paper,
    author = {Filippo Maria Bianchi and others},
    title = {Snow
avalanche segmentation in SAR images with fully convolutional neural networks. },
    journal = {IEEE},
    year = {2021},
    publisher = {IEEE}
}

@article{paper1,
    author = {Alexander Kirillov and others},
    title =  {Segment anything.},
    journal = {IEEE},
    year = {2023},
    publisher = {IEEE}
}

@article{paper2,
    author = {Kathrin Lisa Kapper, Thomas Goelles and others},
    title =  {Automated snow avalanche
monitoring for Austria: State of the
art and roadmap for future work.},
    journal = {Frontiers},
    year = {2023},
    publisher = {Frontiers}
}

@article{paper3,
    author = {Jakob Abermann and others},
    title =  {A large wet snow avalanche cycle in West Greenland 
quantified using remote sensing and in situ observations.},
    journal = {Springer},
    year = {2019},
    publisher = {Springer}
}

@article{paper5,
    author = {HuiHui Dong and others},
    title = {A Multiscale Self-Attention Deep Clustering for
Change Detection in SAR Images},
    journal = {IEEE},
    year = {2022},
    publisher = {IEEE}
}

@article{paper6,
    author = {Jonathan Long and others},
    title = {Fully Convolutional Networks for Semantic Segmentation},
    journal = {arXiv},
    year = {2015},
    publisher = {arXiv}
}

@article{paper9,
    author = {Edward Hu and others},
    title = {LORA: LOW-RANK ADAPTATION OF LARGE LANGUAGE MODELS},
    journal = {arXiv},
    year = {2021},
    publisher = {arXiv}
}

@article{paper11,
    author = {Tal Shaharabany and others},
    title = {AutoSAM: Adapting SAM to Medical Images by
Overloading the Prompt Encoder},
    journal = {arXiv},
    year = {2023},
    publisher = {arXiv}
}

@article{pu2025classwise,
    author = {Xinyang Pu and others},
    title = {ClassWise-SAM-Adapter: Parameter Efficient
Fine-tuning Adapts Segment Anything to SAR
Domain for Semantic Segmentation},
    journal = {IEEE},
    year = {2024},
    publisher = {IEEE}
}

@article{paper13,
    author = {Yun Zhang and others},
    title = {FLOOD AREA SEGMENTATION BY SAM BASED ON SAR DATA AND DEM ASSISTANCE},
    journal = {IGARSS},
    year = {2024},
    publisher = {IGARSS}
}

@article{paper14,
    author = {Lei Ke and others},
    title = {Segment Anything in High Quality},
    journal = {NeurIPS},
    year = {2023},
    publisher = {NeurIPS}
}

@article{paper15,
    author = {Mingjin Zhang, Yuchun Wang and others},
    title = {IRSAM: Advancing Segment Anything Model for
Infrared Small Target Detection},
    journal = {arXiv},
    year = {2024},
    publisher = {arXiv}
}

@article{paper16,
    author = {Junde Wu and others},
    title = {Medical SAM Adapter: Adapting Segment Anything Model for Medical Image Segmentation},
    journal = {arXiv},
    year = {2023},
    publisher = {arXiv}
}

@article{paper17,
    author = {Jun Ma and others},
    title = {Segment anything in medical images},
    journal = {Nature communications},
    year = {2024},
    publisher = {Nature communications}
}

@article{paper18,
    author = {Di Wang and others},
    title = {SAMRS: Scaling-up Remote Sensing Segmentation
Dataset with Segment Anything Mode},
    journal = {NeurIPS},
    year = {2023},
    publisher = {NeurIPS}
}

@article{xiao2024segment,
    author = {Aoran Xiao and others},
    title = {Segment Anything with Multiple Modalities},
    journal = {arXiv},
    year = {2024},
    publisher = {arXiv}
}

@article{paper20,
    author = {Zhiyuan Yan and others},
    title = {RingMo-SAM: A Foundation Model for Segment
Anything in Multimodal Remote-Sensing Images},
    journal = {IEEE},
    year = {2023},
    publisher = {IEEE}
}

@article{paper22,
    author = {Weijie Li and others},
    title = {Predicting Gradient is Better: Exploring Self-Supervised Learning for SAR ATR with a Joint-Embedding Predictive Architecture},
    journal = {Elsevier},
    year = {2024},
    publisher = {Elsevier}
}

@article{paper24,
    author = {Zhiyuan Yan and others},
    title = {Self-supervised training strategies for SAR image despeckling with deep neural networks},
    journal = {Eusar},
    year = {2022},
    publisher = {Eusar}
}

@article{paper31,
    author = {Mathilde Caron and others},
    title = {Emerging Properties in Self-Supervised Vision Transformers},
    journal = {arXiv},
    year = {2021},
    publisher = {arXiv}
}

@article{paper33,
    author = {Kaiming He and others},
    title = {Masked Autoencoders Are Scalable Vision Learners},
    journal = {arXiv},
    year = {2021},
    publisher = {arXiv}
}

@article{paper46,
    author = {JONG-SEN LEE},
    title = {Digital Image Enhancement and Noise Filtering
by Use of Local Statistics},
    journal = {IEEE},
    year = {1980},
    publisher = {IEEE}
}

@article{paper48,
    author = {Lasse Espeholt, Shreya Agrawal and others},
    title = {Deep learning for twelve hour precipitation
forecasts},
    journal = {nature communications},
    year = {2022},
    publisher = {nature communications}
}

@article{paper49,
    author = {Muhammad Waseem Ashraf, Waqas Sultani and others},
    title = {Dogfight: Detecting Drones from Drones Videos},
    journal = {arXiv},
    year = {2021},
    publisher = {arXiv}
}

@article{paper50,
    author = {RONGSHENG DONG , XIAOQUAN PAN AND FENGYING LI},
    title = {DenseU-Net-Based Semantic Segmentation of Small Objects in Urban Remote Sensing Images},
    journal = {IEEEAccess},
    year = {2019},
    publisher = {IEEEAccess}
}

@article{paper53,
    author = {Olaf Ronneberger and Philipp Fischer and Thomas Brox},
    title = {U-Net: Convolutional Networks for Biomedical
Image Segmentation},
    journal = {arXiv},
    year = {2015},
    publisher = {arXiv}
}

@article{paper54,
    author = {Enze Xie and Wenhai Wang and Zhiding Yu},
    title = {SegFormer: Simple and Efficient Design for Semantic
Segmentation with Transformers},
    journal = {arXiv},
    year = {2021},
    publisher = {arXiv}
}

@article{lin2017focal,
  title={Focal loss for dense object detection},
  author={Lin, Tsung-Yi and Goyal, Priya and Girshick, Ross and He, Kaiming and Doll{\'a}r, Piotr},
  journal={Proceedings of the IEEE international conference on computer vision},
  pages={2980--2988},
  year={2017}
}

@article{bianchi2026snow,
  title={Snow avalanches},
  author={Bianchi, Filippo Maria and Grahn, Jakob},
  journal={Data-Driven Earth Observation for Disaster Management},
  pages={69--88},
  year={2026},
  publisher={Elsevier}
}

@article{vickers2016method,
  title={A method for automated snow avalanche debris detection through use of synthetic aperture radar (SAR) imaging},
  author={Vickers, H and Eckerstorfer, M and Malnes, E and Larsen, Y and Hindberg, H},
  journal={Earth and Space Science},
  volume={3},
  number={11},
  pages={446--462},
  year={2016},
  publisher={Wiley Online Library}
}

@article{eckerstorfer2016remote,
  title={Remote sensing of snow avalanches: Recent advances, potential, and limitations},
  author={Eckerstorfer, Markus and B{\"u}hler, Yves and Frauenfelder, Regula and Malnes, Eirik},
  journal={Cold Regions Science and Technology},
  volume={121},
  pages={126--140},
  year={2016},
  publisher={Elsevier}
}

@article{eckerstorfer2015manual,
  title={Manual detection of snow avalanche debris using high-resolution Radarsat-2 SAR images},
  author={Eckerstorfer, Markus and Malnes, Eirik},
  journal={Cold Regions Science and Technology},
  volume={120},
  pages={205--218},
  year={2015},
  publisher={Elsevier}
}

@article{eckerstorfer2019near,
  title={Near-real time automatic snow avalanche activity monitoring system using Sentinel-1 SAR data in Norway},
  author={Eckerstorfer, Markus and Vickers, Hannah and Malnes, Eirik and Grahn, Jakob},
  journal={Remote Sensing},
  volume={11},
  number={23},
  pages={2863},
  year={2019},
  publisher={MDPI}
}

@article{eckerstorfer2017complete,
  title={A complete snow avalanche activity record from a Norwegian forecasting region using Sentinel-1 satellite-radar data},
  author={Eckerstorfer, Markus and Malnes, Eirik and M{\"u}ller, Karsten},
  journal={Cold regions science and technology},
  volume={144},
  pages={39--51},
  year={2017},
  publisher={Elsevier}
}

@article{grahn2024data,
  title={Data-driven avalanche forecasting using weather and satellite data},
  author={Grahn, Jakob and Bianchi, Filippo Maria and M{\"u}ller, Karsten and Malnes, Eirik},
  journal = {International Snow Science Workshop (ISSW) Proceedings},
  year={2024},
  publisher={Montana State University}
}

@inproceedings{
dosovitskiy2021an,
title={An Image is Worth 16x16 Words: Transformers for Image Recognition at Scale},
author={Alexey Dosovitskiy and Lucas Beyer and Alexander Kolesnikov and Dirk Weissenborn and Xiaohua Zhai and Thomas Unterthiner and Mostafa Dehghani and Matthias Minderer and Georg Heigold and Sylvain Gelly and Jakob Uszkoreit and Neil Houlsby},
booktitle={International Conference on Learning Representations},
year={2021},
url={https://openreview.net/forum?id=YicbFdNTTy}
}

@article{bianchi2020large,
  title={Large-scale detection and categorization of oil spills from SAR images with deep learning},
  author={Bianchi, Filippo Maria and Espeseth, Martine M and Borch, Nj{\aa}l},
  journal={Remote Sensing},
  volume={12},
  number={14},
  pages={2260},
  year={2020},
  publisher={MDPI}
}

@article{amitrano2024flood,
  title={Flood detection with SAR: A review of techniques and datasets},
  author={Amitrano, Donato and Di Martino, Gerardo and Di Simone, Alessio and Imperatore, Pasquale},
  journal={Remote Sensing},
  volume={16},
  number={4},
  pages={656},
  year={2024},
  publisher={MDPI}
}

@article{loshchilov2017decoupled,
  title={Decoupled weight decay regularization},
  author={Loshchilov, Ilya and Hutter, Frank},
  journal={arXiv preprint arXiv:1711.05101},
  year={2017}
}

@article{hochreiter1997long,
  title={Long short-term memory},
  author={Hochreiter, Sepp and Schmidhuber, J{\"u}rgen},
  journal={Neural computation},
  volume={9},
  number={8},
  pages={1735--1780},
  year={1997},
  publisher={MIT press}
}

\end{document}